%% file: main.tex
\documentclass[10pt,twocolumn,letterpaper]{article}

\usepackage[pagenumbers]{iccv} % To force page numbers, e.g. for an arXiv version

\definecolor{iccvblue}{rgb}{0.21,0.49,0.74}
\usepackage[pagebackref,breaklinks,colorlinks,allcolors=iccvblue]{hyperref}
\usepackage{colortbl, xcolor, multirow, array, booktabs}
\usepackage{bm}

\def\paperID{****} % *** Enter the Paper ID here
\def\confName{ICCV}
\def\confYear{2025}

\title{Sequential Gaussian Avatars with Hierarchical Motion Context}

\author{Wangze Xu\textsuperscript{1}\footnotemark[1] \quad Yifan Zhan\textsuperscript{1,2} \quad Zhihang Zhong\textsuperscript{1}\textsuperscript{†} \quad Xiao Sun\textsuperscript{1}\textsuperscript{†}\\
\textsuperscript{1}Shanghai Artificial Intelligence Laboratory \quad \textsuperscript{2}The University of Tokyo \\
}

\begin{document}

\twocolumn[{
\renewcommand\twocolumn[1][]{#1}
\maketitle
\centering
\vspace{-3ex}
\includegraphics[width=0.95\linewidth]{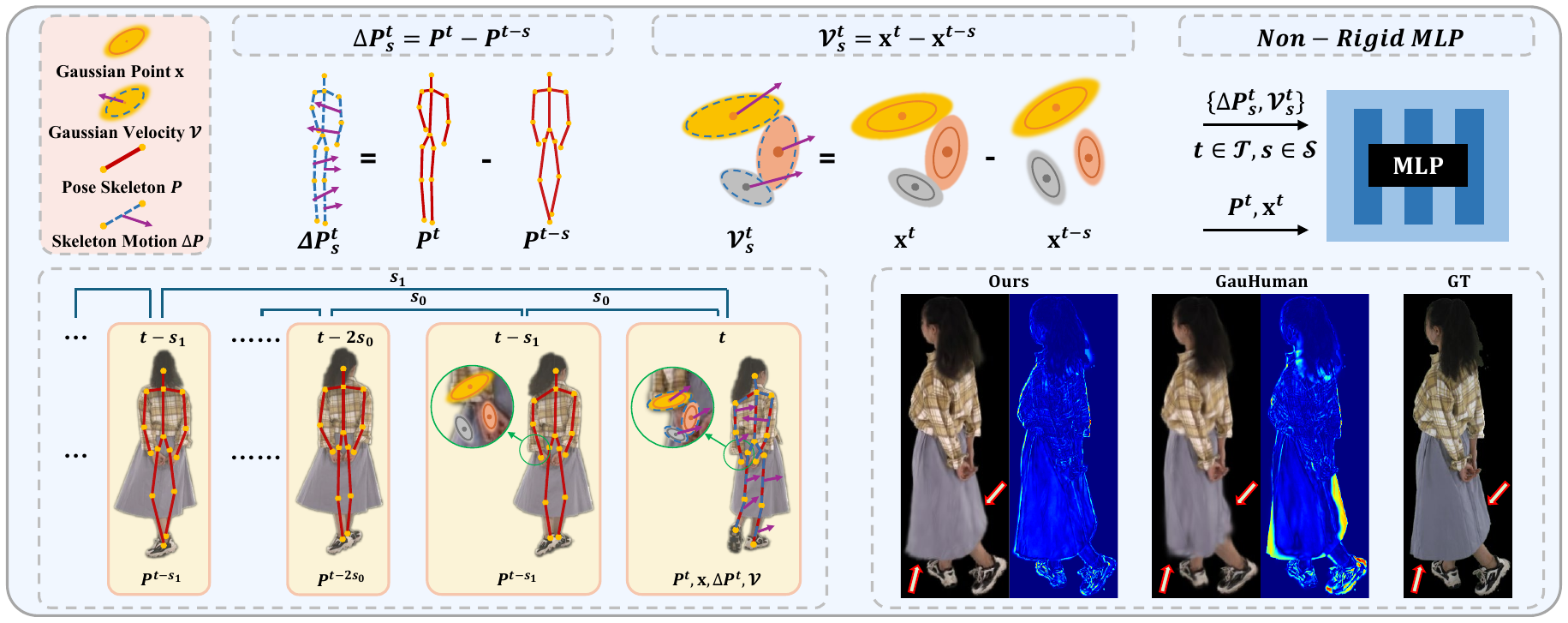}
\captionof{figure}{\textbf{Illustration of the Hierarchical Motion Context Design.} We model the motion-dependent appearance variations with both coarse skeleton condition $\Delta P$ and fine-grained point-wise velocity condition $\mathcal{V}$. $\Delta P$ describes the overall human skeleton motion, which is derived from the difference between the human poses at adjacent frames. $\mathcal{V}$ is the point-wise velocity that indicates finer-grained motion in local regions. To achieve robust deformation, we propose Spatio-temporal Multi-scale Sampling, which samples the overall motion trend and inter-frame details via diverse time intervals $s$ for $\Delta P$ and $\mathcal{V}$. }
\label{fig: teaser}
\vspace{3ex}
}]

\renewcommand{\thefootnote}{\fnsymbol{footnote}}
\footnotetext[1]{This work was done during the author's internship at Shanghai Artificial Intelligence Laboratory. \textsuperscript{†} denotes co-corresponding authors.}

\input{sec/0_abstract}    
\input{sec/1_intro}
\input{sec/2_relatedwork}
\input{sec/3_preliminary}
\input{sec/4_method}
\input{sec/5_experiments}

\input{sec/6_conclusion}

\clearpage
\input{sec/7_acknowledgment}

{
    \small
    \bibliographystyle{ieeenat_fullname}
    \bibliography{main}
}

\input{sec/X_suppl}

\end{document}

%% file: sec/0_abstract.tex
\begin{abstract}
The emergence of neural rendering has significantly advanced the rendering quality of 3D human avatars, with the recently popular 3DGS technique enabling real-time performance. However, SMPL-driven 3DGS human avatars still struggle to capture fine appearance details due to the complex mapping from pose to appearance during fitting. In this paper, we propose SeqAvatar, which excavates the explicit 3DGS representation to better model human avatars based on a hierarchical motion context. Specifically, we utilize a coarse-to-fine motion conditions that incorporate both the overall human skeleton and fine-grained vertex motions for non-rigid deformation. To enhance the robustness of the proposed motion conditions, we adopt a spatio-temporal multi-scale sampling strategy to hierarchically integrate more motion clues to model human avatars. Extensive experiments demonstrate that our method significantly outperforms 3DGS-based approaches and renders human avatars orders of magnitude faster than the latest NeRF-based models that incorporate temporal context, all while delivering performance that is at least comparable or even superior. Project page: \url{https://zezeaaa.github.io/projects/SeqAvatar/}
\end{abstract}

%% file: sec/1_intro.tex
\section{Introduction}
\label{sec:intro}
Recent research on digital humans has highlighted the efficiency of 3D Gaussian Splatting (3DGS)~\cite{kerbl3Dgaussians}, demonstrating its capability for high-quality and real-time rendering.
By defining T-pose human Gaussians in a canonical space and employing specific warping techniques, we can render human avatars from diverse perspectives and in any pose. The canonical Gaussian primitives and warping weights are then jointly optimized under supervision from video inputs.

Animatable human avatar reconstruction is primarily hindered by limited modeling of non-rigid warping, forcing a trade-off between rendering quality and animation capability. Current methods ~\cite{hu2024gauhuman, qian20243dgs} incorporate SMPL(-X) pose priors ~\cite{loper2015smpl, SMPL-X:2019} to guide motion for each observation and rely on Linear Blend Skinning (LBS) for warping. Although these approaches are effective for pose-driven animation, they often struggle with per-frame non-rigid warping in scenarios involving complex garments.

We experimentally find that current human pose conditions~\cite{hu2024gauhuman, qian20243dgs} do not fully capture the complex one-to-many mapping from pose to appearance, particularly for 3DGS-based approaches~\cite{kerbl3Dgaussians}. These methods~\cite{weng2022humannerf,qian20243dgs} rely on the spatial information of a template body in each frame's human pose to predict non-rigid deformations. However, they often overlook local details, such as garment deformations far from the skeleton, and cannot resolve cases where the same pose corresponds to different appearances during complex motions. Furthermore, although previous NeRF-based~\cite{mildenhall2020nerf} method~\cite{chen2024within} has attempted to model motion sequences using human pose residuals, the inherently global nature of the pose sequence limits the ability to capture finer motion details. Naive pose sequence modeling does not fully account for the explicit characteristics of 3DGS.

In this paper, we introduce a hierarchical motion context condition for 3DGS-based human avatar modeling to address the limitations of relying solely on human pose, which provides only limited global skeletal information. Our approach improves the ability of 3DGS-based methods to accurately capture the complex relationship between human pose and appearance in challenging scenarios.
Specifically, we design a coarse-to-fine motion condition that incorporates both overall skeletal movements and fine-grained point-wise motions. Leveraging the explicit nature of Gaussian primitives, this condition seamlessly integrates into 3DGS-based methods, enabling more precise predictions of complex non-rigid deformations. To further enhance robustness, we propose a spatio-temporal multi-scale sampling strategy for constructing the motion context with a larger receptive field. Spatially, we consider the motion states of neighboring points within the local region of each Gaussian primitive to obtain more stable motion embeddings. Temporally, we capture human motion patterns across multiple time scales, combining long-term trends with fine-grained inter-frame motion details. This improves the model’s generalization to complex human movements.

To summarize, our key contributions are as follows:
\begin{itemize}[noitemsep,topsep=0pt]
  \item [1)] We propose a novel hierarchical motion condition that integrates coarse-to-fine human motion, combining global skeletal poses with localized vertex residuals to enhance non-rigid deformation prediction.
  \item [2)] We introduce a spatiotemporal multiscale sampling strategy that expands the receptive field of hierarchical motion context, improving generalization to complex motions.
  \item [3)] Experiments on the I3D-Human~\cite{chen2024within}, DNA-Rendering~\cite{cheng2023dna}, and ZJU-MoCap~\cite{peng2021neural} datasets show the effectiveness of the proposed \textbf{SeqAvatar}, which is capable of modeling details of the human body in complex motions.
\end{itemize}

%% file: sec/2_relatedwork.tex
\section{Related Work}
\label{sec:relatedwork}
\noindent\textbf{Neural Rendering.} Neural rendering techniques have brought significant progress to human reconstruction and rendering. In particular, Neural Radiance Fields (NeRF)~\cite{mildenhall2020nerf} introduces an implicit scene representation that models color and density using multilayer perceptrons, delivering photorealistic rendering results. More recently, point-based 3D Gaussian Splatting \cite{kerbl3Dgaussians} utilizes 3D Gaussians to represent scenes explicitly, achieving real-time high-quality rendering. Building upon NeRF and 3DGS, subsequent works have advanced neural rendering across various dimensions, \textit{e.g.}, enhancing visual fidelity~\cite{barron2021mip, mipnerf360}, improving sparse-view reconstruction quality~\cite{dsnerf, wang2023sparsenerf, zhu2024fsgs, xu2024mvpgs, peng2024structure},  enabling pose-free optimization~\cite{wang2021nerf,lin2021barf,truong2023sparf, shen2024disentangled}, modeling dynamic scenes~\cite{pumarola2021d,fridovich2023k,li2022neural,park2021nerfies, wu2025swift4d, yan2025instant}, and accelerating training and inference~\cite{garbin2021fastnerf,muller2022instant,fridovich2022plenoxels,sun2022direct,chen2022tensorf,liu2020neural}.
These developments also benefit human avatar modeling, which demands high-quality and efficient rendering.

\noindent\textbf{SMPL(-X)-Based Neural Human Modeling.} The SMPL(-X) family~\cite{loper2015smpl, SMPL-X:2019} provides a parametric representation of the human body by decomposing it into pose-related and shape-related components using 3D mesh scans and principal component analysis (PCA). Its \emph{pose blend shapes} enables body deformation through joint-wise pose blending, offering an efficient and compact way for animation. SMPL(-X) has thus become a cornerstone in human body modeling and animation, with many methods~\cite{dong2021fast, shuai2022multinb} estimating its parameters directly from 2D inputs. 
Recent neural human reconstruction methods integrate SMPL(-X) with implicit representations such as NeRF~\cite{mildenhall2020nerf} and 3DGS~\cite{kerbl3Dgaussians} to enable animatable avatars with high rendering quality. Different from purely 2D image animation methods~\cite{xu2024magicanimate, wang2024unianimate, tu2025stableanimator,niu2025anicrafter}, these methods typically register the input data to a canonical T-pose space and use linear blend skinning (LBS) to transform 3D points into observation space based on SMPL(-X) poses. NeRF-based approaches~\cite{weng2022humannerf, peng2021neural, kwon2021neural, chen2024within, goel2023humans, chen2021animatable, chen2023fast, chen2021snarf, gafni2021dynamic, gao2023neural, geng2023learning, wang2022arah, zhan2024tomie} focus on monocular or multi-view reconstruction. More recently, 3DGS-based methods~\cite{li2024animatable, liu2023animatable, zielonka2023drivable, li2023human101, hu2024gauhuman, qian20243dgs, lei2024gart, hu2024surmo, niu2024bundle, jung2023deformable, hu2024gaussianavatar, li2024gaussianbody, liu2024gea, zheng2024gps, kocabas2024hugs, moreau2024human, jena2023splatarmor, zheng2024physavatar, zhan2025r3} have gained popularity due to their real-time rendering and high fidelity. Some works~\cite{chen2024within, hu2024surmo} further incorporate pose sequences as temporal context, but they suffer from slow rendering and lack fine details of motion caused by simply integrating coarse human pose embedding with NeRF.

\noindent\textbf{Human Rendering with Temporal Embeddings} The earliest modeling of dynamic radiance field~\cite{gao2021dynamic,li2022neural,li2021neural,xian2021space,fridovich2023k,cao2023hexplane,liu2023robust,park2023temporal,jang2022d,gan2023v4d,shao2023tensor4d} can be traced back to temporal embeddings in general scenes, where each frame’s observation is obtained by constructing a canonical space and a time-conditioned deformation field.
Motivated by these, a stream of research~\cite {lin2023im4d, xu20244k4d} focuses on pure rendering quality, employing temporal embeddings instead of human pose to encode each frame of human videos. Although these methods achieve high-quality, temporally continuous rendering, they struggle with the lack of geometric constraints from human pose, making parameterized animation unfeasible.

%% file: sec/3_preliminary.tex
\section{Preliminary}
\textbf{SMPL(-X) Series}~\cite{loper2015smpl, SMPL-X:2019} adopt a parameterized framework to represent human bodies across diverse shapes and poses. 
For each frame, the human mesh is derived by deforming a canonical template mesh based on shape and pose parameters.
Specifically, a 3D point $\mathbf{x}$ on the canonical mesh is warped to the corresponding point on the deformed mesh in the observation space, following
\begin{equation}
    \mathbf{LBS}(\mathbf{x}, {\mathbf{B}_{k}}, \mathbf{\omega}_{k}(\mathbf{x}))=\sum_{k=1}^{K}\mathbf{\omega}_{k}(\mathbf{x})\mathbf{B}_{k}\mathbf{x},
\label{equ: LBS}
\end{equation}
where $K$ is the total bone number and $\mathbf{B}_{k}$ is the transformation matrix for each bone. 
Specifically, $\mathbf{B}_{k}$ consists of a rotation and translation matrix $\mathbf{R}_{k}$ and $\mathbf{b}_{k}$. Here, $\mathbf{R}_{k}$ represents the global rotation of each joint, influenced by the local joint rotations $\bm{r}$, while $\mathbf{b}_{k}$ denotes each joint's translation, determined by the joint positions $\bm{j}$ and the human shape parameters $\beta$. The linear blending weight $\omega^k$ depends on $\mathbf{x}$ and is regressed from comprehensive human meshes.

\noindent\textbf{3D Gaussian Splatting}~\cite{kerbl3Dgaussians} utilizes a set of point-based Gaussian primitives to explicitly represent scenes, enabling real-time and high-quality rendering. 
Each Gaussian primitive is defined by its center position $\mathbf{x} \in \mathbb{R}^{3}$ and covariance matrix $\mathbf{\Sigma} \in \mathbb{R}^{3\times3}$. To ensure positive semi-definiteness and simplify optimization, the covariance matrix $\Sigma$ is decomposed into a rotation matrix $\mathbf{R}$ and a scaling matrix $\mathbf{S}$:
\begin{equation}
    \mathbf{\Sigma} = \mathbf{R}\mathbf{S}\mathbf{S}^{T}\mathbf{R}^{T},
\end{equation}
where $\mathbf{R}$ and $\mathbf{S}$ are derived by a scaling vector $\mathbf{s} \in \mathbb{R}^{3}$ and a quaternion vector $\mathbf{r} \in \mathbb{R}^{4}$ in practice. 
Additionally, each Gaussian primitive is also assigned a color feature $sh \in \mathbb{R}^{k}$ represented by spherical harmonics (SH) and an opacity $\mathbf{\alpha}$ for rendering. 
During the rendering process, a splatting technique~\cite{kerbl3Dgaussians} is used to project each Gaussian primitive onto the 2D image space with a viewing transform $\mathbf{W}$ and the Jacobian $\mathbf{J}$ of the projective transformation's affine approximation. 
The transformed covariance $\mathbf{\Sigma}'$ in the camera's coordinates is 
\begin{equation}
    \mathbf{\Sigma}' = \mathbf{J} \mathbf{W} \mathbf{\Sigma} \mathbf{W}^T \mathbf{J}^T.
\end{equation}

After projection, the pixel color $C$ is computed by blending $N$ ordered Gaussian primitives overlapping at the pixel:
\begin{equation}
    C = \sum_{i \in N} c_{i} \alpha'_{i} \prod_{j=1}^{i-1} (1-\alpha'_{j}),
\end{equation}
where $c_i$ is computed from SH feature $sh$ and $\alpha'_{i}$ is the product of $\alpha_{i}$ and probability density of $i$-th 2D Gaussian.

For optimization, a photometric loss is defined by a combination of $L_1$ and SSIM~\cite{wang2004image} losses:
\begin{equation}
   L_{photo} = \lambda L_1(\hat{I}, I) + (1 - \lambda)(1 - SSIM(\hat{I}, I)),
\end{equation}
where $\hat{I}$ and $I$ denote the rendered and ground-truth images, and $\lambda$ controls the balance between the two terms. 

%% file: sec/4_method.tex
\section{Method}
\label{sec:method}
\begin{figure*}[t]
    \centering
    \includegraphics[width=0.95\linewidth]{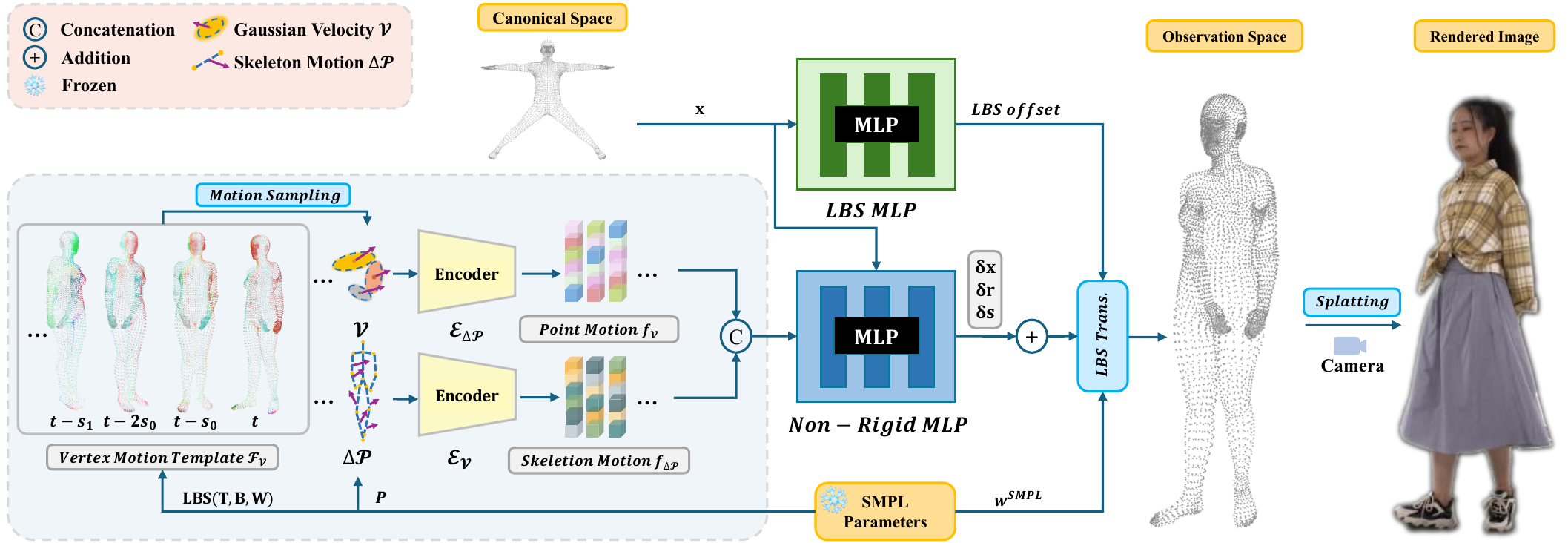}
    \caption{\textbf{Overview of the proposed method.} We first initialize canonical Gaussian positions $\mathbf{x}$ with SMPL template vertexes. For each Gaussian, we derive both coarse skeleton motion condition $f_{\Delta \mathcal{P}}$ and fine-grained vertex motion condition $f_{\mathcal{V}}$ that sampled from the vertex motion template $\mathcal{F}_{\mathcal{V}}$ (points in different colors represent different motions). Based on such hierarchical motion information, we utilize an MLP $\mathcal{E}_{non-rigid}$ to better predict each Gaussian's non-rigid deformation prediction. The non-rigid deformed Gaussians are then warped into observation space via the standard LBS transformation for rendering.}
    \label{fig:framework}
\end{figure*}

Fig. \ref{fig:framework} shows the framework of our method. We use the explicit point-based 3DGS as the representation of the human body. Given a collection of input cameras and images, we optimize a set of Gaussian primitives $\{\mathcal{G}_{i}\}_{i=1}^{i=n}$ to fit the body's shape and appearance. Each Gaussian primitive $\mathcal{G}_{i}$ includes the center position $\mathbf{x}$, scaling vector $\mathbf{s}$, rotation quaternion vector $\mathbf{r}$, color feature $sh$ and opacity $\mathbf{\alpha}$, where $\mathbf{x}$ is initialized from the SMPL template vertices $\{\mathbf{T}_{i}\}_{i=1}^{i=N}$. During the optimization process, we first apply a coarse-to-fine motion context to capture more accurate and fine-grained details (Sec. \ref{subsec:non_rigid_cond}). To mitigate overfitting, we propose Spatio-Temporal Multi-Scale Sampling to obtain more robust point-wise Gaussian motion, which serves as the embedding for non-rigid deformation (Sec.\ref{subsec:STMS}). Next, we adopt Linear Blend Skinning (LBS) to map canonical Gaussian primitives $\mathcal{G}$ to observation space and render images via differentiable splitting (Sec.\ref{subsec:optimization}).

\subsection{Non-Rigid Deformation with Coarse-to-fine Motion Context
}
\label{subsec:non_rigid_cond}
\textbf{Coarse Skeleton Motion.} As illustrated in Fig. \ref{fig: teaser}, we introduce a coarse-to-fine motion context to excavate additional conditions for non-rigid deformation. Typically, most human 3D avatar methods use the current frame pose as a condition to predict non-rigid deformation~\cite{hu2024gauhuman, qian20243dgs}. While this provides spatial information to distinguish the motion of different body parts, it fails to capture temporal motion changes, as discussed in~\cite{chen2024within}. Therefore, given a frame at time $t$, we consider a sequence of regularly interval-sampled frames $\mathcal{T}$ to model inter-frame body motion variations:
\begin{equation}
    \label{eq:define_sequence}
    \mathcal{T}=\{t-s, t-2s, ..., t-Ls\},
\end{equation}
where $L$ is the sequence length and $s$ is the time interval. The coarse motion of body skeletons at each time step $t \in \mathcal{T}$ can be derived by calculating the difference of poses between adjacent frames as illustrated in Fig. \ref{fig: teaser}, following
\begin{equation}
    \label{eq:skeleton_delta_motion_encoder}
    \Delta \mathcal{P} = \{\Delta P^{t} = \delta (P^{t}, P^{t-s}) | t \in \mathcal{T}\},
\end{equation}
where $P \in \mathbb{R}^{K \times 3}$ is the body pose and $\delta$ denotes the difference $\Delta \mathcal{P}$ between two poses in axis-angle form. We employ an MLP \boldsymbol{$\mathcal{E}_{\Delta \mathcal{P}}$} to encode the sequential skeleton motion $\Delta \mathcal{P} \in \mathbb{R}^{L \times K \times 3}$, flattened along the temporal dimension, into a skeleton motion embedding $f_{\Delta \mathcal{P}} \in \mathbb{R}^{32}$, 
\begin{equation}
    \label{eq:vertex_delta_motion_encoder}
    f_{\Delta \mathcal{P}} = \mathcal{E}_{\Delta \mathcal{P}}(\Delta \mathcal{P}),
\end{equation}
which serves as a condition for the subsequent non-rigid deformation. Further details are provided in the supplementary material.

\noindent\textbf{Fine Vertex Motion.} Compared to previous implicit NeRF-based methods, the point-based 3DGS representation enables us to explore more fine-grained temporal motion information. We derive a point-wise velocity vector $v_{i}$ for each Gaussian primitive $\mathcal{G}_{i}$ to model the fine-grained localized body motion, which is beyond the scope of pose skeleton motion $\Delta \mathcal{P}$. 
By measuring the position variations across adjacent time steps, we calculate per-frame velocity
\begin{equation}
    v^{t} = \frac{\mathbf{x_{o}}^{t} - \mathbf{x_{o}}^{t-s}}{s},
\end{equation}
where $\mathbf{x_{o}}$ denotes the warped coordinates in observation space. However, since the positions of Gaussian primitives are updated during optimization, it is not stable to obtain $v_{i}$ based on such a dynamic variable. Moreover, transforming the canonical Gaussians into the observation space requires non-rigid deformation first, which leads to a circular dependency and conflicts with the current velocity design. To this end, we introduce a motion template field \boldsymbol{$\mathcal{F}_{\mathbf{V}}=\{\mathbf{V}_{i}\}_{i=1}^{i=N}$} that stores the velocity of each SMPL vertex, and $v_{i}$ for each Gaussian primitive is sampled from \boldsymbol{$\mathcal{F}_{\mathbf{V}}$} based on the distance between query points and vertexes in this template. Specifically, given a time step $t$ and the corresponding body pose $P^{t}=\{(\mathbf{R}_{k}^{t}, \mathbf{b}_{k}^{t}) | k \in K\}$, the SMPL template vertexes in $\mathbf{T}$ can be warped into observation space with the template skinning weights $\mathbf{W}$ and standard linear blend skinning, following
\begin{equation}
    \mathbf{T_{o}}^{t} = \mathbf{LBS}(\mathbf{T}, \mathbf{B}^{t}, \mathbf{W}),
\end{equation}
where $\mathbf{B}^{t}$ is the transformation matrix derived from rotation and translation matrix $\mathbf{R}^{t}$ and $\mathbf{b}^{t}$. $\mathbf{W}$ is the template SMPL LBS weights.
The velocity $\mathbf{V}$ of each template vertexes $\mathbf{T}$ can be further derived as
\begin{equation}
    \mathbf{V}^{t} = \frac{\mathbf{T_{o}}^{t} - \mathbf{T_{o}}^{t-s}}{s}.
\end{equation}
We then sample each Gaussian's velocity from this motion template field \boldsymbol{$\mathcal{F}_{\mathbf{V}}$}, which is discussed in Sec. \ref{subsec:STMS}. 

\subsection{Spatio-temporal Multi-scale Sampling (STMS)}
\label{subsec:STMS}
To enhance the robustness of human motion conditions, we propose modeling both local region motion information of the human body and motion patterns across different temporal windows to mitigate overfitting and improve generalization by capturing more comprehensive motion dynamics.

In the spatial dimension, we sample the $\tau$ nearest template to model the body's local region motion more robustly. Specifically, for each canonical Gaussian primitive $\mathcal{G}_{i}$, the $\tau$ nearest vertexes' velocities are sampled as input to an MLP $\mathcal{E}_{knn}$ to learn a local point-wise motion embedding
\begin{equation}
    \label{eq:knn_velocity}
    e_{i}^{t}=\mathcal{E}_{knn}(\{\mathbf{V}_{j}^{t}\}), \quad j \in \mathbf{KNN}(\mathbf{T}, \mathbf{x}_{i}),
\end{equation}
where $\mathbf{KNN}(\mathbf{T}, \mathbf{x}_{i})$ denotes the $\tau$ nearest SMPL template vertexes of the canonical Gaussian position $\mathbf{x}_{i}$. Similar to Eq. \ref{eq:skeleton_delta_motion_encoder}, we then apply an MLP \boldsymbol{$\mathcal{E}_{\mathcal{V}}$} to encode each Gaussian primitive's sequential motion embedding $\mathcal{V}=\{e^{t} | t \in \mathcal{T}\}$ into a point-wise sequential condition $f_{\mathcal{V}} \in \mathbb{R}^{96}$
\begin{equation}
    \label{eq:vertex_delta_motion_encoder}
    f_{\mathcal{V}} = \mathcal{E}_{\mathcal{V}}(\mathcal{V}).
\end{equation}

In the temporal dimension, to capture both the overall motion trend and the inter-frame motion details, we adopt a multi-scale sequence sampling strategy. 
In detail, we sample a series of sequences at several progressively increasing intervals to get human motion information across different temporal windows
\begin{equation}
    \label{eq:different_sample_scale}
    \mathcal{S} = \{s=s_{0}+i\Delta s\}_{i=0}^{i=m} ,
\end{equation}
where $\Delta s$ is the increasing rate of sampling interval $s$. We then input the multi-scale sampled sequence motions into the skeleton motion encoder \boldsymbol{$\mathcal{E}_{\Delta \mathcal{P}}$} and the point-wise motion encoder \boldsymbol{$\mathcal{E}_{\mathcal{V}}$} to obtain hierarchical temporal motion embeddings. Therefore, Eq. \ref{eq:skeleton_delta_motion_encoder} and Eq. \ref{eq:vertex_delta_motion_encoder} can be revised as
\begin{align}
    f_{\Delta \mathcal{P}} &= \mathcal{E}_{\Delta \mathcal{P}}(\{\Delta \mathcal{P}_{s}\}) , \quad s \in \mathcal{S} \\ f_{\mathcal{V}} &= \mathcal{E}_{\mathcal{V}}(\{\mathcal{V}_{s}\}), \quad s \in \mathcal{S}.
\end{align}
In practice, we concatenate all skeleton motion conditions, $\Delta \mathcal{P}_{s}$ for $s \in \mathcal{S}$, and localized vertex motion conditions, $\mathcal{V}_{s}$ for $s \in \mathcal{S}$, across different sampling scales. These are then input to \boldsymbol{$\mathcal{E}_{\Delta \mathcal{P}}$} and \boldsymbol{$\mathcal{E}_{\mathcal{V}}$}, respectively. More details are shown in the supplementary material.

\noindent\textbf{Non-Rigid Deformation.} Given the coarse skeleton motion $f_{\Delta \mathcal{P}}=\mathcal{E}_{\Delta \mathcal{P}}(\{\Delta \mathcal{P}_{s}\})$ condition and fine point-wise velocity condition $f_{\mathcal{V}}=\mathcal{E}_{\mathcal{V}}(\{\mathcal{V}_{s}\})$, we utilize an MLP to predict each Gaussian's non-rigid deformation as
\begin{equation}
    \label{eq:non_rigid_mlp}
    \delta \mathbf{x}, \delta \mathbf{s}, \delta \mathbf{r} = \mathcal{E}_{non-rigid}(\mathbf{x}, P, f_{\Delta \mathcal{P}}, f_{v}).
\end{equation}
The deformed canonical Gaussian $\mathcal{G}'$ is
\begin{align}
    \mathbf{x'} &= \mathbf{x} + \delta \mathbf{x} ,\\
    \mathbf{s'} &= \mathbf{s} + \delta \mathbf{s} ,\\
    \mathbf{r'} &= \mathbf{r} \cdot \delta \mathbf{r} ,
\end{align}
where $\cdot$ denotes the multiplication of two quaternions.

\subsection{Optimization}
\label{subsec:optimization}
\textbf{Rigid Deformation.} We utilize the standard $\mathbf{LBS}$ operation to map the non-rigid deformed Gaussians $\mathcal{G}_{o}$ into the observation space, following
\begin{align}
    \mathbf{x_{o}} &= \mathbf{LBS}(\mathbf{x'}, \mathbf{B}, \mathbf{\omega}) , \\
    \mathbf{R_{o}} &= \sum_{k=1}^{K}\mathbf{\omega}_{k}(\mathbf{x'})\mathbf{B}_{k} \mathbf{R}',
\end{align}
where $\mathbf{R}'$ is the rotation matrix derived from the non-rigid deformed Gaussian's rotation quaternion $\mathbf{r}'$, and $\mathbf{R_{o}}$ represents the rotation matrix in observation space. Following the previous method~\cite{hu2024gauhuman}, we utilize an MLP $\mathcal{E}_{lbs}$ to predict the LBS weight offsets for each query canonical point and update the sampled weights from the nearest SMPL vertex as
\begin{equation}
    \mathbf{\omega}_{k}(\mathbf{x}) = \mathbf{\omega}_{k}^{SMPL}(\mathbf{x}) + \mathcal{E}_{lbs}(\mathbf{x}).
\end{equation}
Similar to~\cite{weng2022humannerf, chen2024within, hu2024gauhuman}, we introduce a pose refinement MLP $\mathcal{E}_{pose}$ to refine the pose estimate from SMPL for a better fit to the human body.

\noindent\textbf{Loss Function.} With the transformed Gaussian primitives $\mathcal{G}_{o}$ in observation space, we apply the standard splitting~\cite{kerbl3Dgaussians} to render images as
\begin{equation}
    I = \mathbf{Splatting}(\mathbf{x_{o}}, \mathbf{R_{o}}, s', \alpha, sh).
\end{equation}
During the optimization process, we employ a combination of loss functions as supervision, summarized as
\begin{equation}
    \mathcal{L} = \lambda_{1} \mathcal{L}_{color} + \lambda_{2} \mathcal{L}_{ssim} + \lambda_{3} \mathcal{L}_{lpips} + \mathcal{L}_{mask},
\end{equation}
where $\lambda$ is used to balance the weight of different losses. $\mathcal{L}_{mask}$~\cite{hu2024gauhuman} is an $L_{2}$ loss between the rendered $\alpha$ and human body mask. Similar to~\cite{qian20243dgs}, we apply $\mathcal{L}_{isopos}$ and $\mathcal{L}_{isocov}$ to control the Gaussian primitive's position and covariance. Details are shown in the supplementary material.

%% file: sec/5_experiments.tex
\section{Experiments}
\label{sec:experiments}

\subsection{Datasets}
\textbf{DNA-Rendering}~\cite{cheng2023dna} is a challenging human-centric rendering dataset. 
It contains diverse scenes from everyday life to professional occasions. 
We use 6 sequences with loose-fitting garments (1\_0206\_04, 2\_0007\_04, 2\_0019\_10, 2\_0044\_11, 2\_0051\_09, 2\_0813\_05) for experiments. 
We adopt 24 views for training and 6 views for testing. 

\noindent\textbf{I3D-Human}~\cite{chen2024within} contains scenes closer to daily life, which comprises multi-view video frames of humans with loose clothing and complex movements. We conduct experiments on 4 sequences and use 4-5 views for training and the rest for testing. All training and testing data are retained to be the same in comparisons following Dyco's~\cite {chen2024within} data split. 

\input{table/quantitative_results_dna}

\begin{figure*}[h]
    \centering
    \includegraphics[width=0.9\linewidth]{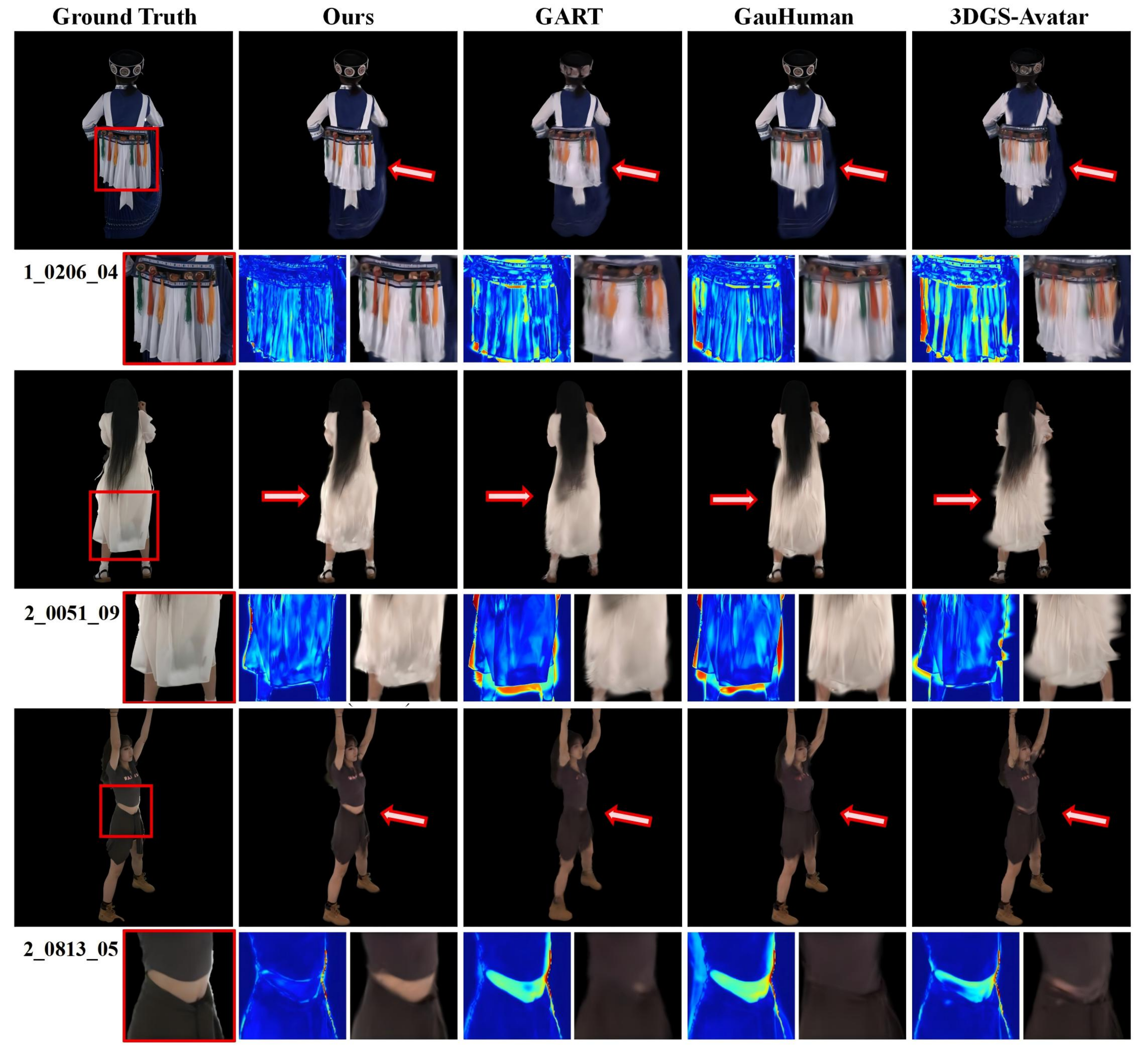}
    \caption{\textbf{Novel View Qualitative Results on DNA-Rendering.} We zoom into the local region and compute the error maps compared with ground truth images. The results show that our method achieves competitive results on both the overall and local region qualities. }
    \label{fig:quality_dna}
\end{figure*}

\input{table/quantitative_results_i3d}

\begin{figure*}[t]
    \centering
    \includegraphics[width=0.95\linewidth]{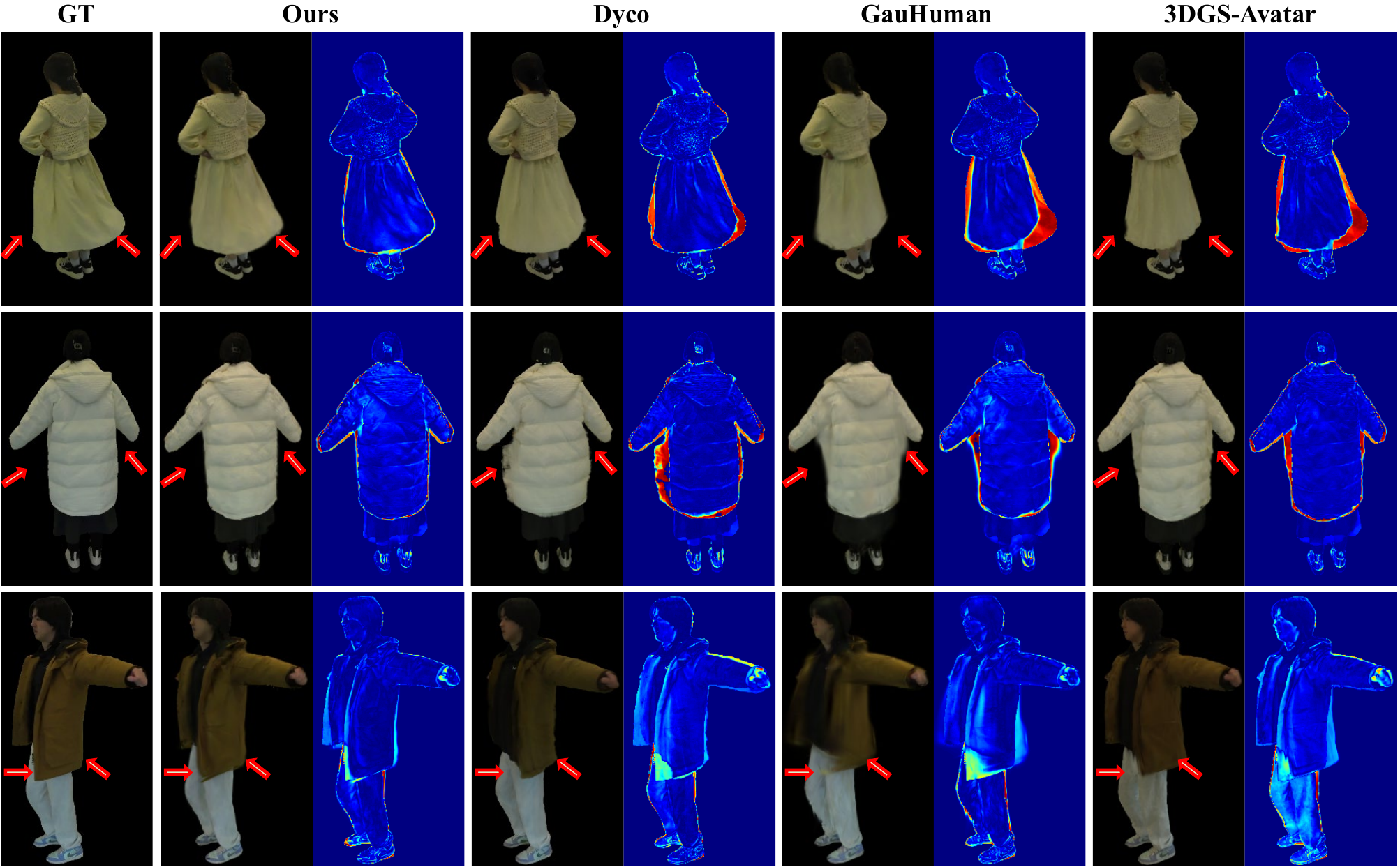}
    \caption{\textbf{Novel Pose Qualitative Results on I3D-Human.} We compare the proposed method with previous SOTA approaches. The rendered images and error maps demonstrate our robustness on novel pose rendering.}
    \label{fig:supp_novelpose_i3d}
\end{figure*}

\subsection{Baselines and Metrics}

We compare against the state-of-the-art human modeling methods including Dyco~\cite{chen2024within}, 3DGS-Avatar~\cite{qian20243dgs}, GART~\cite{lei2024gart}, and GauHuman~\cite{hu2024gauhuman}. Dyco is an implicit NeRF-based method that adopts pose sequence information for temporal modeling. We compare performance with Dyco to showcase the better efficacy of our fine-grained motion condition design. 3DGS-Avatar, GART, and GauHuman are explicit 3DGS-based methods, and all of them are originally designed for monocular inputs. For fairness, we extend them to multi-view input under the same settings. We report three key metrics: peak signal-to-noise ratio (PSNR), structural similarity index measure (SSIM)~\cite{wang2004image}, and learned perceptual image patch similarity (LPIPS)~\cite{zhang2018unreasonable}.

\subsection{Comparison}

\begin{figure*}[htpb]
    \centering
    \includegraphics[width=0.95\linewidth]{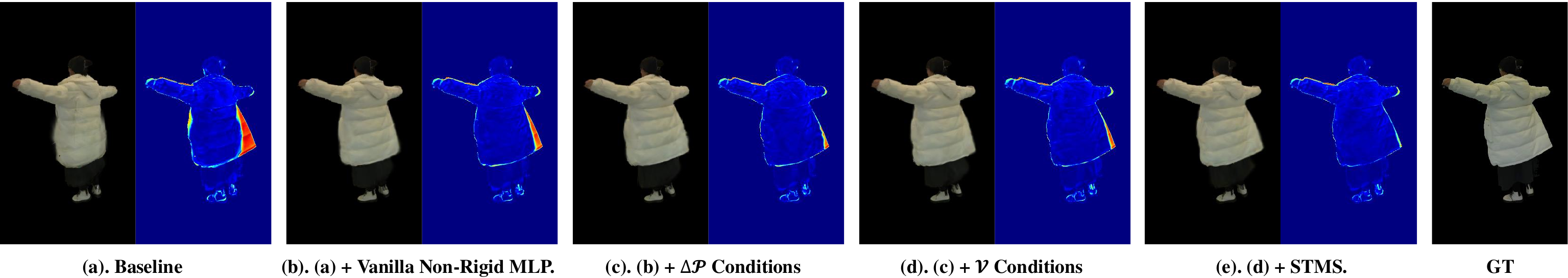}
    \caption{\textbf{Qualitative Ablation Results.} We compare the visual influence of different components on ID2\_1 scene of I3D-Human dataset.}
    \label{fig:ablation_vis}
\end{figure*}

\begin{figure}[h]
    \centering
    \includegraphics[width=1.0\linewidth]{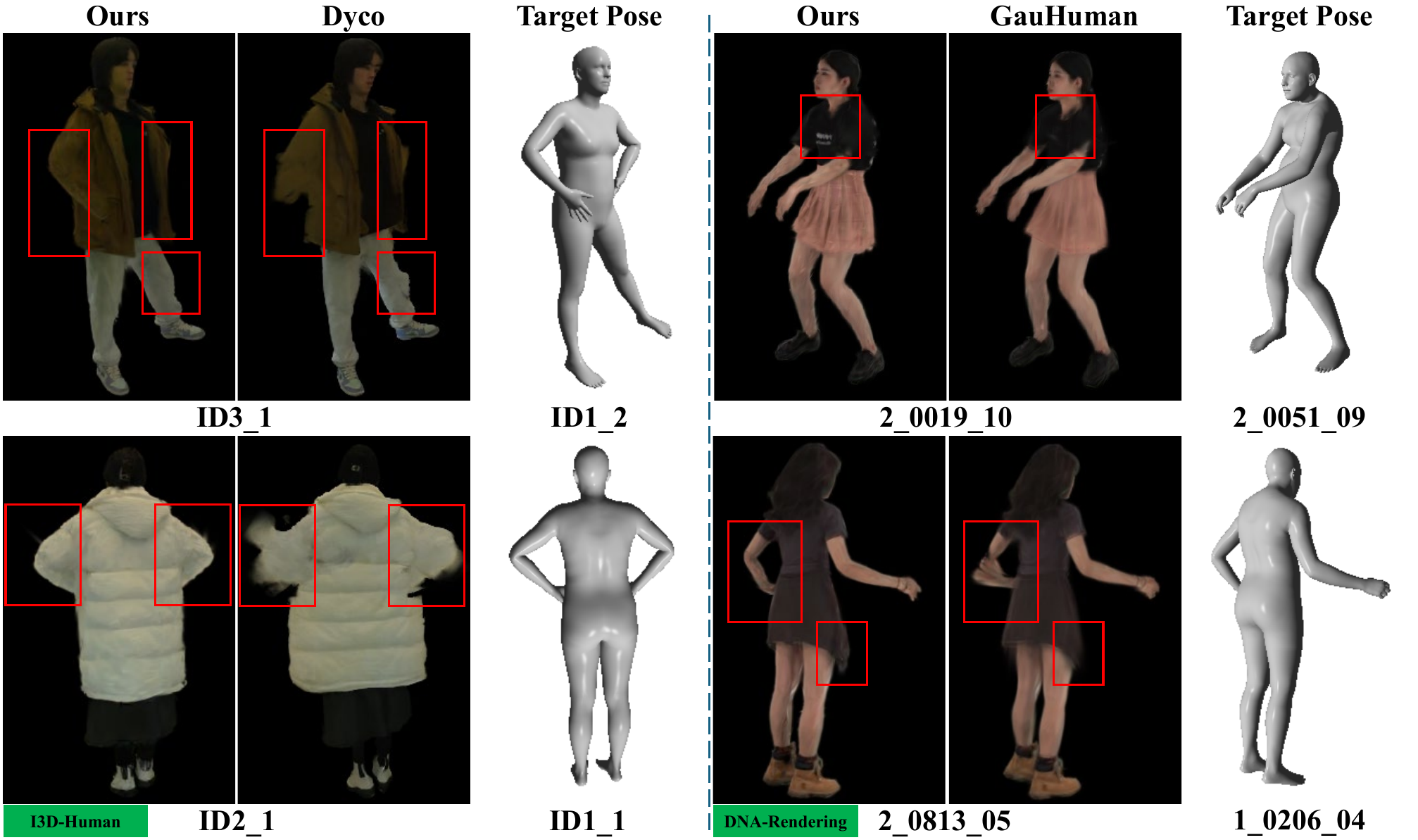}
    \caption{\textbf{Out-of-distribution Pose Animation.}}
    \label{fig:qualitative_ood_pose}
\end{figure}

\textbf{Comparison on DNA-Rendering.} Tab. \ref{tab:quantitative_dna_res} shows the quantitative results on DNA-Rendering dataset. The proposed method outperforms previous state-of-the-art human modeling methods across all metrics. To demonstrate the visual improvement, we compare the quality of rendering images in Fig. \ref{fig:quality_dna}. Previous methods, which lack the temporal motion information for non-rigid deformations, are unable to predict the movement of Gaussian primitives properly, resulting in blurred renderings in regions with complex textures (scene 1\_0206\_04 in Fig. \ref{fig:quality_dna}). Besides, the results in Fig. \ref{fig:quality_dna} show that the compared methods fail to capture the proper shape variation caused by movement (e.g., the white dress in scene 2\_0051\_09). Although these methods achieve a smooth appearance in these areas, the error maps show that they fail to match the actual shape closely. We leverage the hierarchical motion context to predict Gaussian deformations, allowing our method to capture detailed appearance variations caused by human motions more accurately.

\noindent\textbf{Comparison on I3D-Human.} The quantitative results in Tab. \ref{tab:quantitative_i3d_res} show that our method surpasses the previous SOTA method on most metrics, demonstrating our competitive performance on the I3D-Human dataset. Previous NeRF-based Dyco~\cite{chen2024within} utilizes pose variation as a condition to model human motions. However, it is limited to capturing only overall body motions and lacks the capacity for finer-grained modeling in local regions. For example, in ID1\_1 and ID2\_1 shown in Fig. \ref{fig:quality_i3d} (in the Supp. Mat.), our method renders results that accurately reflect the true motion, while Dyco fails to model the information in regions far from the human body skeleton. Tab. \ref{tab:supp_novelpose_quantitative_i3d_res} reports the quantitative results of novel pose rendering on the I3D-Human dataset. Our method achieves best performance among 3DGS-based human modeling methods and also enables real-time rendering ($\sim$ 45 FPS on I3D-Human) compared to NeRF-based Dyco ($\sim$ 0.7 FPS). Fig. \ref{fig:supp_novelpose_i3d} shows the qualitative comparisons of novel pose rendering. Different from Dyco~\cite{chen2024within}, which relies solely on joint motions as conditions, our method leverages the motion state of each Gaussian primitive to predict deformations, enabling more flexible modeling of motions in regions distant from the human body.

To evaluate the generalizability, we also conduct experiments on ZJU-MoCap~\cite{peng2021neural}. Please refer to the supplementary materials for more details.

\noindent\textbf{Out-of-Distribution Poses.} To evaluate the generalization ability to novel poses, we conduct experiments using animations with large pose variations. Specifically, we use a model trained on one sequence to render target poses sampled from other unseen sequences, as different sequences in DNA-Rendering and I3D-Human contain distinct motion patterns. As illustrated in Fig.~\ref{fig:qualitative_ood_pose}, the \textit{Target Pose} is drawn from an unseen sequence exhibiting different human motions. The results suggest that our method is capable of generalizing to these significant pose changes, even though such poses were not observed during training.

\input{table/quantitative_reulsts_i3d_novelpose}

\subsection{Ablation Study}
In this section, we ablate the influence of various components in our methods on I3D-Human dataset. Tab. \ref{tab:ablation} shows the average metrics of 4 sequences under different settings.

\noindent\textbf{Skeleton motion condition $\Delta \mathcal{P}$} denotes the overall human body movement. Compared to predicting non-rigid deformation only with the Gaussian position, incorporating this additional global temporal information allows our method to better model the overall human shape and achieve higher rendering quality, as demonstrated in Tab. \ref{tab:ablation} (c).

\noindent\textbf{Fine-grained point-wise motion condition $\mathcal{V}$} describes the local region's variation in a more detailed manner. Such point-based motion information captures the relationship between motion and appearance variations in areas beyond the human skeleton. Tab. \ref{tab:ablation} (d) and Fig. \ref{fig:ablation_vis} (d) show that our method achieves higher rendering details in local regions.

\noindent\textbf{Spatio-temporal Multi-scale Sampling (STMS)} is designed to enhance the robustness of human motion conditions by incorporating local region motion information and motion patterns across different temporal windows. With such a hierarchical motion context embedding design, our method is able to predict each Gaussian's deformation more robustly, leading to better rendering quality as shown in Tab. \ref{tab:ablation} (e) and Fig. \ref{fig:ablation_vis} (e). Please refer to the supplementary materials for more detailed ablation experiments.

\input{table/ablation_i3d}

%% file: table/quantitative_results_dna.tex
\begin{table*}[h]
\centering
\caption{\textbf{Quantitative Results on DNA-Rendering Dataset.}
We report the performance of novel view rendering. Our method outperforms previous state-of-the-art human modeling methods. We mark the \textbf{\colorbox[RGB]{255,179,179}{best}} and the \textbf{\colorbox[RGB]{255,217,179}{second best}} methods in cells. LPIPS*=LPIPS $\times 10^{3}$.}
\resizebox{1.0 \linewidth}{!}{
\begin{tabular}{l|ccc|ccc|ccc|ccc|ccc|ccc}
\toprule
Subject: & \multicolumn{3}{c}{1\underline{\hspace{0.7em}}0206\underline{\hspace{0.7em}}04} & \multicolumn{3}{c}{2\underline{\hspace{0.7em}}0007\underline{\hspace{0.7em}}04} & \multicolumn{3}{c}{2\underline{\hspace{0.7em}}0019\underline{\hspace{0.7em}}10} & \multicolumn{3}{c}{2\underline{\hspace{0.7em}}0044\underline{\hspace{0.7em}}11} & \multicolumn{3}{c}{2\underline{\hspace{0.7em}}0051\underline{\hspace{0.7em}}09} & \multicolumn{3}{c}{2\underline{\hspace{0.7em}}0813\underline{\hspace{0.7em}}05} \\ 
Metric: & PSNR$\uparrow$ & SSIM$\uparrow$ & LPIPS*$\downarrow$ & PSNR$\uparrow$ & SSIM$\uparrow$ & LPIPS*$\downarrow$ & PSNR$\uparrow$ & SSIM$\uparrow$ & LPIPS*$\downarrow$ & PSNR$\uparrow$ & SSIM$\uparrow$ & LPIPS*$\downarrow$ & PSNR$\uparrow$ & SSIM$\uparrow$ & LPIPS*$\downarrow$ & PSNR$\uparrow$ & SSIM$\uparrow$ & LPIPS*$\downarrow$ \\ 
\midrule
3DGS-Avatar~\cite{qian20243dgs}
& 26.64 & 0.9439 & \cellcolor{orange!25}50.09
& 28.06 & 0.9492 & 56.70
& \cellcolor{orange!25}32.06 & \cellcolor{orange!25}0.9710 & \cellcolor{orange!25}31.12
& 28.23 & 0.9512 & 37.15
& 25.04 & 0.9534 & 42.22
& 31.75 & 0.9701 & 31.30 \\ 
GART~\cite{lei2024gart} 
& 27.19 & 0.9454 & 56.47
& \cellcolor{orange!25}28.22 & \cellcolor{orange!25}0.9525 & \cellcolor{orange!25}56.46
& 31.13 & 0.9686 & 35.16 
& 28.90 & 0.9561 & 41.27 
& \cellcolor{orange!25}26.47 & \cellcolor{orange!25}0.9628 & 42.21 
& 32.06 & 0.9729 & 35.74 \\ 
GauHuman~\cite{hu2024gauhuman}
& \cellcolor{orange!25}27.96 & \cellcolor{orange!25}0.9500 & 50.42
& 27.93 & 0.9496 & 56.93
& 31.63 & 0.9698 & 32.82
& \cellcolor{orange!25}30.38 & \cellcolor{orange!25}0.9641 & \cellcolor{orange!25}34.19
& 25.99 & 0.9599 & \cellcolor{orange!25}42.00
& \cellcolor{orange!25}33.43 & \cellcolor{orange!25}0.9764 & \cellcolor{orange!25}29.42 \\ 
\hline
Ours
& \cellcolor{red!25}30.81 & \cellcolor{red!25}0.9649 & \cellcolor{red!25}39.47
& \cellcolor{red!25}29.63 & \cellcolor{red!25}0.9566 & \cellcolor{red!25}43.72
& \cellcolor{red!25}34.97 & \cellcolor{red!25}0.9777 & \cellcolor{red!25}23.53
& \cellcolor{red!25}32.62 & \cellcolor{red!25}0.9747 & \cellcolor{red!25}23.83
& \cellcolor{red!25}28.45 & \cellcolor{red!25}0.9685 & \cellcolor{red!25}34.06
& \cellcolor{red!25}35.81 & \cellcolor{red!25}0.9843 & \cellcolor{red!25}20.83 \\ 
\bottomrule
\end{tabular}
}
\label{tab:quantitative_dna_res}
\end{table*}

%% file: table/quantitative_results_i3d.tex
\begin{table*}[h]
\centering
\caption{\textbf{Novel View Rendering Quantitative Results on I3D-Human.}}
\resizebox{0.9 \linewidth}{!}{
\begin{tabular}{l|ccc|ccc|ccc|ccc}
\toprule
Subject: & \multicolumn{3}{c}{ID1\underline{\hspace{0.7em}}1} & \multicolumn{3}{c}{ID1\underline{\hspace{0.7em}}2} & \multicolumn{3}{c}{ID2\underline{\hspace{0.7em}}1} & \multicolumn{3}{c}{ID3\underline{\hspace{0.7em}}1} \\ 
Metric: & PSNR$\uparrow$ & SSIM$\uparrow$ & LPIPS*$\downarrow$ & PSNR$\uparrow$ & SSIM$\uparrow$ & LPIPS*$\downarrow$ & PSNR$\uparrow$ & SSIM$\uparrow$ & LPIPS*$\downarrow$ & PSNR$\uparrow$ & SSIM$\uparrow$ & LPIPS*$\downarrow$ \\ 
\midrule
3DGS-Avatar~\cite{qian20243dgs} & 29.78 & 0.9600 & 31.40 & 31.29 & 0.9627 & 28.09 & \cellcolor{orange!25}29.63 & 0.9604 & 37.33 & \cellcolor{orange!25}32.73 & 0.9602 & 39.47 \\ 
Dyco~\cite{chen2024within}
& \cellcolor{orange!25}30.81 & \cellcolor{orange!25}0.9617 & \cellcolor{orange!25}28.29 
& \cellcolor{orange!25}31.45 & \cellcolor{orange!25}0.9628 & \cellcolor{red!25}25.75 
& 29.41 & \cellcolor{orange!25}0.9618 & \cellcolor{orange!25}33.26 
& 32.50 & \cellcolor{orange!25}0.9612 & \cellcolor{orange!25}35.46 \\
GauHuman~\cite{hu2024gauhuman} & 29.47 & 0.9571 & 39.46 & 30.42 & 0.9585 & 38.87 & 28.51 & 0.9546 & 49.46 & 32.10 & 0.9546 & 53.67 \\ 
\hline
Ours
& \cellcolor{red!25}31.81 & \cellcolor{red!25}0.9659 & \cellcolor{red!25}27.03 
& \cellcolor{red!25}31.98 & \cellcolor{red!25}0.9660 & \cellcolor{orange!25}27.84 
& \cellcolor{red!25}31.53 & \cellcolor{red!25}0.9685 & \cellcolor{red!25}30.23 
& \cellcolor{red!25}33.62 & \cellcolor{red!25}0.9651 & \cellcolor{red!25}34.02 \\
\bottomrule
\end{tabular}
}
\label{tab:quantitative_i3d_res}
\end{table*}

%% file: table/quantitative_reulsts_i3d_novelpose.tex
\begin{table}[htpb]
    \centering
    \caption{\textbf{Metrics of Novel Pose Renderings on I3D-Human.}}
    \resizebox{0.7 \linewidth}{!}{
        \begin{tabular}{l | c c c}
            \toprule
            Methods & PSNR $\uparrow$   & SSIM $\uparrow$   & LPIPS* $\downarrow$ \\ \hline
            Dyco~\cite{chen2024within} & \cellcolor{orange!25}30.09 & \cellcolor{orange!25}0.9569 & \cellcolor{red!25}35.32   \\ 
            3DGS-Avatar~\cite{qian20243dgs} & 29.54 & 0.9555 & 38.59   \\ 
            GauHuman~\cite{hu2024gauhuman} & 29.31 & 0.9526 & 48.29   \\
            \hline
            Ours & \cellcolor{red!25}30.28 & \cellcolor{red!25}0.9583 & \cellcolor{orange!25}36.07   \\ 
            \bottomrule
        \end{tabular}
    }
    \label{tab:supp_novelpose_quantitative_i3d_res}
\end{table}

%% file: table/ablation_i3d.tex
\begin{table}[htpb]
    \centering
    \caption{\textbf{Quantitative Ablation Results on I3D-Human.} (a) denotes experiments without non-rigid deformation. (b) only utilizes Gaussian positions $\mathbf{x}$ and the current frame pose $P$ as the conditions for non-rigid MLP.}
    \resizebox{1.0 \linewidth}{!}{
        \begin{tabular}{l | c c c}
            \toprule
            Methods & PSNR $\uparrow$   & SSIM $\uparrow$   & LPIPS* $\downarrow$ \\ \hline
            (a) Baseline    & 29.76 & 0.9569 & 38.35   \\ 
            \hline
            (b) (a)+Vanilla Non-Rigid MLP+$\mathcal{P}$ Conditions.    & 31.05 & 0.9617 & 34.35   \\ 
            (c) (b)+$\Delta \mathcal{P}$ Conditions.   & 31.89 & 0.9645 & 32.17   \\ 
            (d) (c)+ $\mathcal{V}$ Conditions.    & 32.01 & 0.9651 & 31.23   \\ 
            (e) (d)+STMS. (Ours)    & \textbf{32.24} & \textbf{0.9664} & \textbf{29.78}   \\ 
            \bottomrule
        \end{tabular}
    }
    \label{tab:ablation}
\end{table}

%% file: sec/6_conclusion.tex
\section{Limitations and Conclusion}
\label{sec: Limitations and Conclusion}
\noindent\textbf{Limitations.} The Gaussian-based representation used in our method may introduce slight blur artifacts during rendering, whereas NeRF’s ray-based volumetric integration tends to produce sharper results. Moreover, our local velocity cues are derived from the coarse SMPL model rather than dense surface tracking~\cite{zheng2024physavatar}, which may limit the accuracy of fine-scale garment deformation. Addressing these limitations remains an open challenge for future work.

\noindent\textbf{Conclusion.} In this paper, we propose a 3DGS-based framework that integrates hierarchical motion context for 3D human modeling. Specifically, the non-rigid deformation for the Gaussian primitive is learned based on the global human skeleton variations and fine-grained Gaussian's point-wise motions. To capture each Gaussian primitive's motion information more robustly, we introduce a spatio-temporal multi-scale sampling strategy to incorporate both local region motion features and motion patterns across different temporal intervals. Through the above design, the proposed method achieves state-of-the-art rendering of human avatars with complex garments and motions.

%% file: sec/7_acknowledgment.tex
\section*{Acknowledgment}
This research was supported by Shanghai Artificial Intelligence Laboratory.

%% file: sec/X_suppl.tex
\clearpage
\setcounter{page}{1}
\maketitlesupplementary

\section{Details of Skeleton Motion $\Delta P$}
The pose $P$ is represented as the rotation relative to the parent node of all joints:
\begin{equation}
    P = \{\theta_{1}, \theta_{2}, ..., \theta_{K}\},
\end{equation}
where $\theta_{j} \in \mathbb{R}^{3}$ describes the $j$-th joint's relative rotation with its parent in axis-angle form~\cite{loper2015smpl}. We follow Dyco~\cite{chen2024within} to derive each joint's rotation variation $\Delta \theta$ between adjacent frames as its skeleton motion representation. Specifically, the rotation of $j$-th joint in matrix form can be obtained according to \textit{Rodrigues formula}~\cite{loper2015smpl}:
\begin{equation}
    \mathbf{R}_{j} = Rodrigues(\theta_{j}),
\end{equation}
where $\mathbf{R}_{j}$ is a 3 $\times$ 3 rotation matrix. Given 2 consecutive poses $P^{t}$ and $P^{t-s}$, the temporal variation of $j$-th joint's rotation is computed as a relative transformation matrix $\Delta \mathbf{R}_{j}^{t}$:
\begin{equation}
    \Delta \mathbf{R}_{j}^{t} = \mathbf{R}_{j}^{t}{\mathbf{R}_{j}^{t-s}}^{-1}.
\end{equation}
Then we convert the 3 $\times$ 3 matrix $\Delta \mathbf{R}_{j}^{t}$ back to the axis-angle form $\Delta \theta_{j}^{t} \in \mathbb{R}^{3}$. Now, the skeleton motion at time $t$ is
\begin{equation}
    \Delta P^{t} = \{\Delta\theta_{1}^{t}, \Delta\theta_{2}^{t}, ..., \Delta\theta_{K}^{t}\}, 
\end{equation}
which is simplified as $\Delta P^{t}=\delta(P^{t},P^{t-s})$ in Eq. (\ref{eq:skeleton_delta_motion_encoder}).

\section{Datasets} 
\textbf{I3D-Human Dataset~\cite{chen2024within}}. We use images at a resolution of 512~$\times$512 for experiments and follow Dyco\cite{chen2024within} for both training and testing splits.

\noindent\textbf{DNA-Rendering Dataset~\cite{cheng2023dna}.} We use images with a resolution of 512~$\times$~612 for training and 750~$\times$~1024 for testing in our experiments. For the temporal split, we select 100 consecutive time steps (pose IDs 0–99) and use the multi-view images of all selected steps for training. During testing, we evaluate on a total of 20 time steps sampled every 5 steps from this training range, using multi-view images from held-out camera views. For the view split, we uniformly sample 24 views from camera IDs 0–48 for training and 6 additional views uniformly sampled from camera IDs 48–60 for novel view evaluation.

\noindent\textbf{ZJU-MoCap~\cite{peng2021neural}}. We use images with resolution 512 $\times$ 512 for experiments and follow 3DGS-Avatar\cite{qian20243dgs} for both training and testing splits.

\section{Motion Embeddings for Non-Rigid MLP} 
The embeddings $f_{\Delta \mathcal{P}}$ and $f_{\mathcal{V}}$ in Eq. (\ref{eq:non_rigid_mlp}) describe the skeleton and local region motions, respectively. 

For each frame, all Gaussian primitives $\{\mathcal{G}_{i}\}$ share the same skeleton motion embedding $f_{\Delta \mathcal{P}} \in \mathbb{R}^{32}$ and we concat it with $\mathcal{G}_{i}$'s point-wise motion embedding ${f_{\mathcal{V}}}_{i} \in \mathbb{R}^{96}$ as a new condition $f_{i}' \in \mathbb{R}^{128}$. $f_{i}'$ is then input to an MLP $\mathcal{E}_{non-rigid}$ to predict the corresponding non-rigid deformation. $\mathcal{E}_{non-rigid}$ consists of 3 hidden layers and one output layer, and each hidden layer is followed by a ReLU activation.

\section{Details of Loss Functions}
$\mathcal{L}_{color}$ is the $L_{1}$ loss between the rendered image $I$ and the ground truth $I_{gt}$:
\begin{equation}
    \mathcal{L}_{color} = | I - I_{gt}|.
\end{equation}
We use $\mathcal{L}_{ssim}$ to constrain the structure similarity between the rendered image and the ground truth, which is given by
\begin{equation}
    \mathcal{L}_{ssim} = 1- \text{SSIM}(I,I_{gt}),
\end{equation}
where $\text{SSIM}(\cdot)$ is the SSIM metric. Additionally, we use the LPIPS loss to ensure the perceptual similarity:
\begin{equation}
    \mathcal{L}_{lpips} = \text{LPIPS}(I,I_{gt}),
\end{equation}
where $\text{LPIPS}(\cdot)$ is the LPIPS metric. Following GauHuman~\cite{hu2024gauhuman} and 3DGS-Avatar~\cite{qian20243dgs}, we employ a mask loss to ensure that Gaussian primitives are accurately localized within their designated regions:
\begin{equation}
    \mathcal{L}_{mask} = {||M-M_{gt}||}_{2}
\end{equation}
where $M_{gt}$ is the foreground mask and $M$ is the accumulated $\alpha$ value:
\begin{equation}
    M=\sum \alpha_{i}\prod_{j=1}^{i-1}(1-\alpha_{j}).
\end{equation}

Similar to~\cite{qian20243dgs, prokudin2023dynamic}, we also utilize the as-isometric-as-possible constraint~\cite{kilian2007geometric} to enforce neighboring distance similarity of 3D Gaussian centers and covariance matrices between canonical space and observation space:
\begin{align}
     \mathcal{L}_{isopos}&=\sum_{i=1}^{N}\sum_{j \in \mathbf{KNN}(i)}|d(\mathbf{x}^{i},\mathbf{x}^{j})-d(\mathbf{x_{o}}^{i},\mathbf{x_{o}}^{j})|,
    \\
    \mathcal{L}_{isocov}&=\sum_{i=1}^{N}\sum_{j \in \mathbf{KNN}(i)}|d(\mathbf{\Sigma}^{i},\mathbf{\Sigma}^{j})-d(\mathbf{\Sigma_{o}}^{i},\mathbf{\Sigma_{o}}^{j})|,
\end{align}
where $N$ is the number of Gaussian primitives. $\mathbf{x}, \mathbf{\Sigma}$ and $\mathbf{x_{o}}, \mathbf{\Sigma_{o}}$ are the Gaussian primitive's position and covariance matrix in canonical space and observation space, respectively. $\mathbf{KNN}$ denotes the $K=5$ nearest points.

\begin{figure*}[tp]
    \centering
    \includegraphics[width=1.0\linewidth]{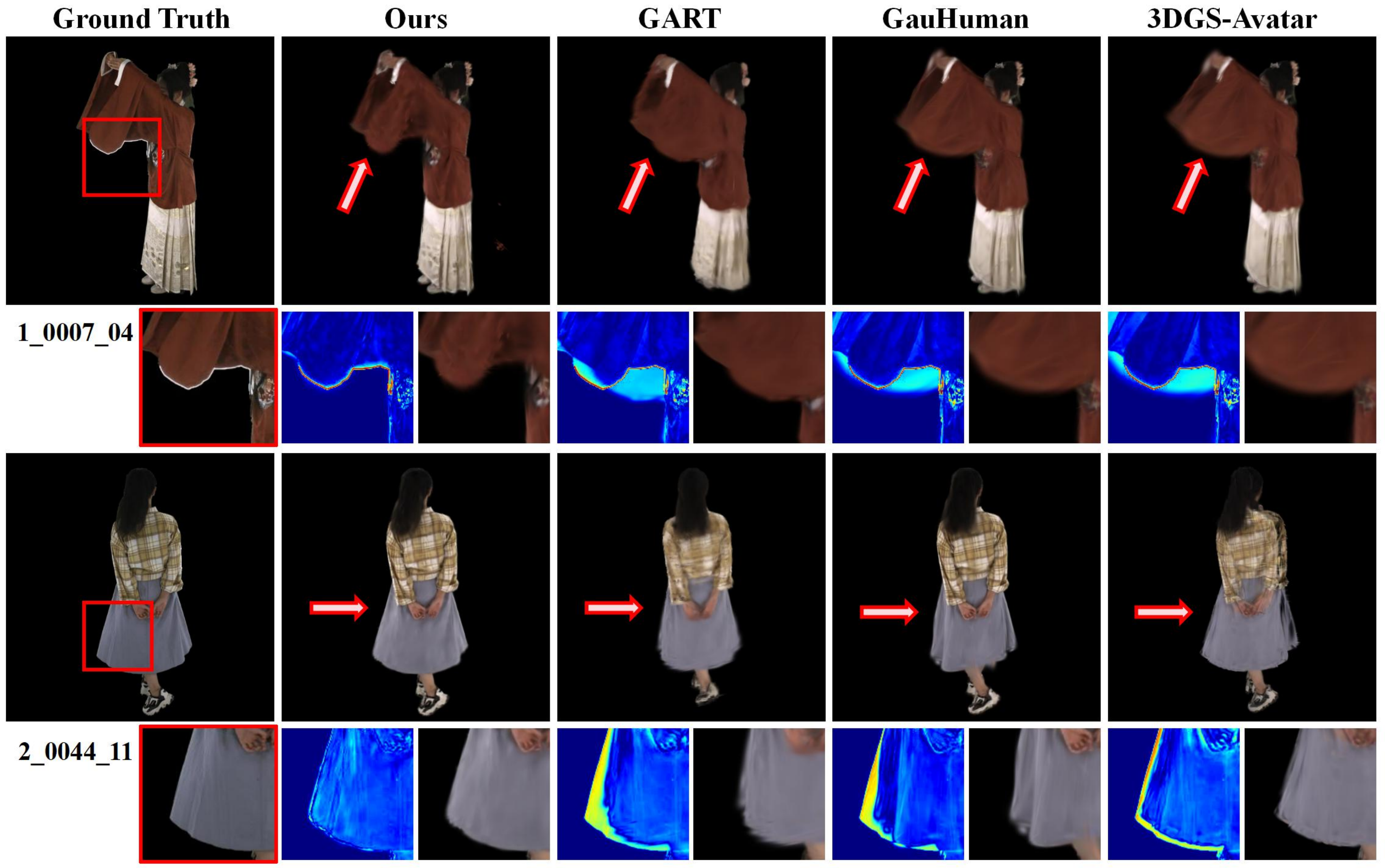}
    \caption{\textbf{Novel View Qualitative Results on DNA-Rendering.} We provide both the complete image and localized error map comparisons against other methods for a more comprehensive visualization.}
    \label{fig:supp_novelview_dna}
\end{figure*}

\section{Details of Optimization}
In our experiments, the number $\lvert \mathcal{S} \rvert$ (in Eq. (\ref{eq:different_sample_scale})) of sequence sampled in different scales is set to 3, and each sequence length $L$ (in Eq. (\ref{eq:define_sequence})) is set to 8. We use the $\tau=8$ nearest SMPL template vertexes' velocities to get each Gaussian primitive's local velocity embedding $e_{i}$ (in Eq. (\ref{eq:knn_velocity})). Additionally, we utilize the adaptive densification in ~\cite{kerbl3Dgaussians} to control the number of Gaussian primitives. 

For the DNA-Rendering dataset, we use the SMPL-X~\cite{SMPL-X:2019} model with pose dimension $P \in \mathbb{R}^{25\times3}$ in our experiments. We adopt the SMPL-X template with $N=10475$ vertexes as Gaussian initialization. We optimize 30k iterations and set the loss weights $\lambda_{0}=1.0, \lambda_{1}=0.01, \lambda_{2}=0.01$. The initial sampling step and the increasing interval are $s_{0}=1$ and $\Delta s = 2$, respectively. The final sample scales are $\mathcal{S}=\{1, 3, 5\}$ (in Eq. (\ref{eq:different_sample_scale})). For the I3D-Human dataset, we follow Dyco~\cite{chen2024within} to use the SMPL~\cite{loper2015smpl} model in our experiments for fairness. The number of SMPL template vertexes used for Gaussian initialization is $N=6890$ and the corresponding pose dimension is $P \in \mathbb{R}^{23\times3}$. We optimize 15k iterations and set the loss weights $\lambda_{0}=1.0, \lambda_{1}=0.1, \lambda_{2}=0.1$. We set the increasing interval $\Delta s = 9$. The initial sampling step is $s_{0}=24$ and the final sample scales are $\mathcal{S}=\{24, 33, 42\}$ (in Eq. (\ref{eq:different_sample_scale})). Note that the experiment in Tab. \ref{tab:ablation} (d) adopts single sampling step that $\mathcal{S}=\{s_{0}\}$.

\begin{figure*}[htp]
    \centering
    \includegraphics[width=1.0\linewidth]{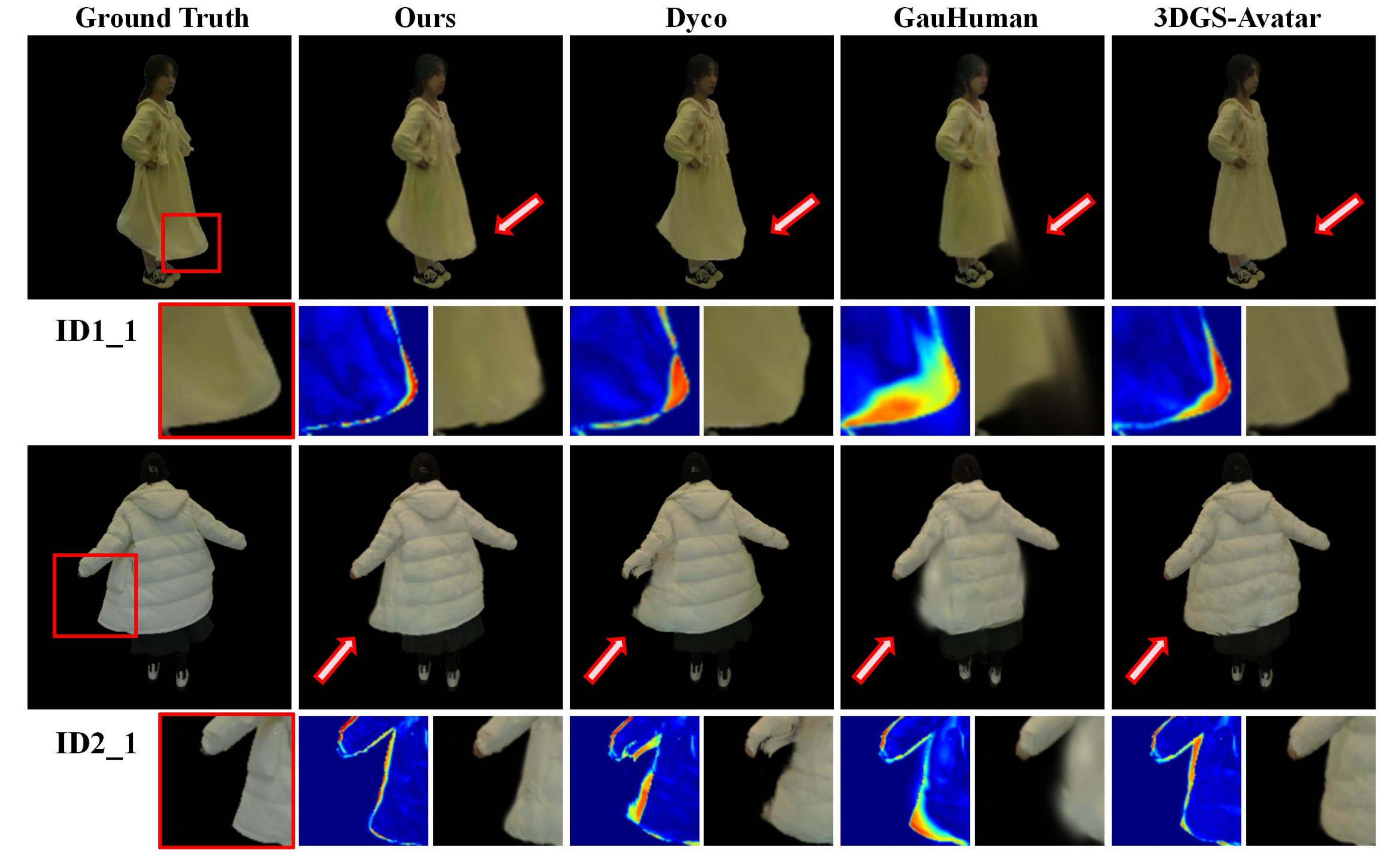}
    \caption{\textbf{Novel View Qualitative Results on I3D-Human.} }
    \label{fig:quality_i3d}
\end{figure*}

\section{Additional Results}
\label{sec:more_results}
\subsection{Experiments on ZJU-MoCap}

\noindent\textbf{ZJU-MoCap}~\cite{peng2021neural} is a common benchmark dataset in human avatar, which is mainly collected under controlled speeds and tight-fitting garments. We use 6 sequences (377, 386, 387, 392, 393, 394) for experiments following the dataset split in 3DGS-Avaar\cite{qian20243dgs}.

\begin{figure}[h]
    \centering
    \includegraphics[width=1.0\linewidth]{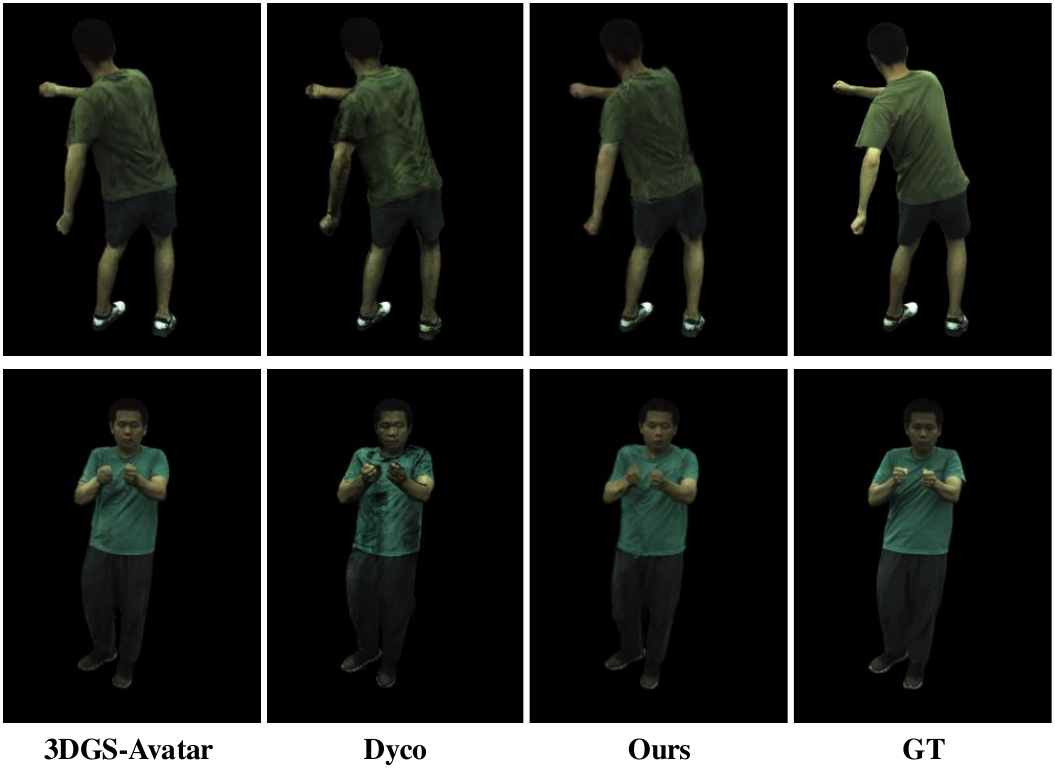}
    \caption{\textbf{Qualitative Results on ZJU-MoCap.}}
    \label{fig:supp_qualitative_zjumocap}
\end{figure}

\noindent\textbf{Comparison on ZJU-MoCap.} Results on DNA-Rendering and I3D-Human demonstrate that our methods achieve better performance in complex cases with loose clothing and uncontrolled speeds. Furthermore, Tab. \ref{tab:quantitative_zjumocap_res} and Fig. \ref{fig:supp_qualitative_zjumocap} show that our method also achieves competitive metrics in simpler controlled scenarios, which verifies the generalizability of the proposed methods. 

\input{table/quantitative_results_zjumocap_ori}

\subsection{Visual Comparisons On DNA-Rendering}

Fig. \ref{fig:supp_novelview_dna} shows additional qualitative comparisons with other SOTA methods on the DNA-Rendering dataset~\cite{cheng2023dna}. Our approach excels in regions where motion causes notable appearance variations, producing results that conform to the true shape more accurately, as exemplified by the sleeve (top row) and the skirt hem (bottom row) in Fig. \ref{fig:supp_novelview_dna}.

\subsection{Video Quality}
We provide rendered videos on the DNA-Rendering dataset~\cite{cheng2023dna}, along with the corresponding error maps similar to Fig.~\ref{fig:quality_dna}. Please refer to the attached videos for comparison.
In the videos, the novel view results are rendered on temporal frames with pose IDs 0–99, using camera viewpoints that are held out during training. 

\section{Additional Ablations}
\label{sec:more_abalation}

\textbf{Different Sample Steps}
Tab. \ref{tab:supp_ablation_delta_steps} shows the quantitative metrics with different sequential sampling steps. The results demonstrate that combining all coarse-to-fine temporal motions as the condition leads to more robust non-rigid deformation and better performance.

\input{table/ablation_multi_step}

\noindent\textbf{Different KNN Sampled Vertexes} Tab. \ref{tab:supp_ablation_knn_num} presents the quantitative results on I3D-Human dataset~\cite{chen2024within} with different numbers of KNN vertexes when deriving the velocity $e_{i}$ of each Gaussian primitive (Eq. \ref{eq:knn_velocity}). It suggests that considering the motion states of all local regions helps obtain a more robust Gaussian primitive's velocity embedding, contributing to enhancing performance.

\input{table/abalation_knn_num}

\section{Computational Cost Analysis}
\input{table/computation_cost_analysis}

We provide computational cost analysis for experiments on ZJU-Mocap (conducted on a single 4090 GPU). Tab. \ref{tab:computational_cost} shows that our approach achieves competitive computational efficiency. The additional computation time for each frame's Vertex Motion Template (multi-scale sampling number $|\mathcal{S}|=3$, sequence length $L=8$) is about 0.09 s, and it only needs to be computed once during preprocessing before training.

%% file: table/quantitative_results_zjumocap_ori.tex
\begin{table}[htpb]
    \centering
    \caption{\textbf{Quantitative Results on ZJU-MoCap.}}
    \resizebox{0.7 \linewidth}{!}{
        \begin{tabular}{l | c c c}
            \toprule
            Methods & PSNR $\uparrow$   & SSIM $\uparrow$   & LPIPS* $\downarrow$ \\ \hline
            Dyco~\cite{chen2024within} & 30.37 & 0.9599 & 29.47   \\ 
            3DGS-Avatar~\cite{qian20243dgs} & 30.59 & 0.9609 & \cellcolor{red!25}27.09   \\ 
            GauHuman~\cite{hu2024gauhuman} & \cellcolor{red!25}31.04 & \cellcolor{red!25}0.9620 & 31.81   \\
            \hline
            Ours & \cellcolor{orange!25}31.02 &\cellcolor{orange!25} 0.9619 & \cellcolor{orange!25}28.89   \\ 
            \bottomrule
        \end{tabular}
    }
    \label{tab:quantitative_zjumocap_res}
\end{table}

%% file: table/ablation_multi_step.tex
\begin{table}[htpb]

    \centering
    
    \caption{\textbf{Impact of Different Sampling Steps on I3D-Human.}}
    \resizebox{0.8 \linewidth}{!}{
    \begin{tabular}{c | c c c}
    \toprule
     Sampling Step $\mathcal{S}$ & PSNR $\uparrow$   & SSIM $\uparrow$   & LPIPS* $\downarrow$ \\ 
    \hline
    $\mathcal{S}=\{24\}$    & 32.01 & 0.9655 & 30.71   \\ 
    $\mathcal{S}=\{33\}$   & 32.09 & 0.9656 & 30.88   \\ 
    $\mathcal{S}=\{42\}$    & 32.11 & 0.9658 & 30.81   \\ 
    \hline
    $\mathcal{S}=\{24, 33, 42\}$    & \textbf{32.24} & \textbf{0.9664} & \textbf{29.78}   \\ 
    \bottomrule
    \end{tabular}
    }
    \label{tab:supp_ablation_delta_steps}

\end{table}

%% file: table/abalation_knn_num.tex
\begin{table}[htpb]

    \centering
    
    \caption{\textbf{Impact of Different KNN Vertexe Number $\tau$.}}
    \resizebox{0.7 \linewidth}{!}{
    \begin{tabular}{c | c c c}
    \toprule
     KNN num. $\tau$ & PSNR $\uparrow$   & SSIM $\uparrow$   & LPIPS* $\downarrow$ \\ 
    \hline
    $\tau=1$    & 32.14 & 0.9659 & 30.41   \\ 
    $\tau=5$   & 32.15 & 0.9660 & 30.32   \\ 
    $\tau=8$    & \textbf{32.24} & 
    \textbf{0.9664} & \textbf{29.79}   \\ 
    $\tau=10$    & 32.20 & 0.9663 & 30.11   \\ 
    \bottomrule
    \end{tabular}
    }
    \label{tab:supp_ablation_knn_num}

\end{table}

%% file: table/computation_cost_analysis.tex
\begin{table}[h]
    \centering
    \caption{\textbf{Computational Efficiency.}}
    \resizebox{0.8 \linewidth}{!}{
    \begin{tabular}{cccc}
        \toprule
         Mehods & Train Time & FPS & Train Mem.  \\
         \midrule
         Dyco & 6 h & 0.7 & 20 GB \\
         3DGS-Avatar & 25 m & 25 & 6 GB \\
         \hline
         Ours ($3k$ $iter$) & 5 m & 62 & 8 GB\\
    \bottomrule
    \end{tabular}
    }
    \label{tab:computational_cost}
\end{table}

%% file: main.bbl
\begin{thebibliography}{81}
\providecommand{\natexlab}[1]{#1}
\providecommand{\url}[1]{\texttt{#1}}
\expandafter\ifx\csname urlstyle\endcsname\relax
  \providecommand{\doi}[1]{doi: #1}\else
  \providecommand{\doi}{doi: \begingroup \urlstyle{rm}\Url}\fi

\bibitem[Barron et~al.(2021)Barron, Mildenhall, Tancik, Hedman, Martin-Brualla, and Srinivasan]{barron2021mip}
Jonathan~T Barron, Ben Mildenhall, Matthew Tancik, Peter Hedman, Ricardo Martin-Brualla, and Pratul~P Srinivasan.
\newblock {Mip-NeRF: A Multiscale Representation for Anti-aliasing Neural Radiance Fields}.
\newblock In \emph{Proceedings of the IEEE/CVF International Conference on Computer Vision}, pages 5855--5864, 2021.

\bibitem[Barron et~al.(2022)Barron, Mildenhall, Verbin, Srinivasan, and Hedman]{mipnerf360}
Jonathan~T Barron, Ben Mildenhall, Dor Verbin, Pratul~P Srinivasan, and Peter Hedman.
\newblock Mip-nerf 360: Unbounded anti-aliased neural radiance fields.
\newblock In \emph{Proceedings of the IEEE/CVF Conference on Computer Vision and Pattern Recognition}, pages 5470--5479, 2022.

\bibitem[Cao and Johnson(2023)]{cao2023hexplane}
Ang Cao and Justin Johnson.
\newblock {Hexplane: A Fast Representation for Dynamic Scenes}.
\newblock In \emph{Proceedings of the IEEE/CVF Conference on Computer Vision and Pattern Recognition}, pages 130--141, 2023.

\bibitem[Chen et~al.(2022)Chen, Xu, Geiger, Yu, and Su]{chen2022tensorf}
Anpei Chen, Zexiang Xu, Andreas Geiger, Jingyi Yu, and Hao Su.
\newblock {Tensorf: Tensorial radiance fields}.
\newblock In \emph{European Conference on Computer Vision}, pages 333--350. Springer, 2022.

\bibitem[Chen et~al.(2021{\natexlab{a}})Chen, Zhang, Kang, Zhe, Bao, Jia, and Lu]{chen2021animatable}
Jianchuan Chen, Ying Zhang, Di Kang, Xuefei Zhe, Linchao Bao, Xu Jia, and Huchuan Lu.
\newblock Animatable neural radiance fields from monocular rgb videos.
\newblock \emph{arXiv preprint arXiv:2106.13629}, 2021{\natexlab{a}}.

\bibitem[Chen et~al.(2021{\natexlab{b}})Chen, Zheng, Black, Hilliges, and Geiger]{chen2021snarf}
Xu Chen, Yufeng Zheng, Michael~J Black, Otmar Hilliges, and Andreas Geiger.
\newblock Snarf: Differentiable forward skinning for animating non-rigid neural implicit shapes.
\newblock In \emph{Proceedings of the IEEE/CVF International Conference on Computer Vision}, pages 11594--11604, 2021{\natexlab{b}}.

\bibitem[Chen et~al.(2023)Chen, Jiang, Song, Rietmann, Geiger, Black, and Hilliges]{chen2023fast}
Xu Chen, Tianjian Jiang, Jie Song, Max Rietmann, Andreas Geiger, Michael~J Black, and Otmar Hilliges.
\newblock Fast-snarf: A fast deformer for articulated neural fields.
\newblock \emph{IEEE Transactions on Pattern Analysis and Machine Intelligence}, 45\penalty0 (10):\penalty0 11796--11809, 2023.

\bibitem[Chen et~al.(2024)Chen, Zhan, Zhong, Wang, Sun, Qiao, and Zheng]{chen2024within}
Yutong Chen, Yifan Zhan, Zhihang Zhong, Wei Wang, Xiao Sun, Yu Qiao, and Yinqiang Zheng.
\newblock Within the dynamic context: Inertia-aware 3d human modeling with pose sequence.
\newblock \emph{arXiv preprint arXiv:2403.19160}, 2024.

\bibitem[Cheng et~al.(2023)Cheng, Chen, Fan, Yin, Chen, Cai, Wang, Gao, Yu, Lin, et~al.]{cheng2023dna}
Wei Cheng, Ruixiang Chen, Siming Fan, Wanqi Yin, Keyu Chen, Zhongang Cai, Jingbo Wang, Yang Gao, Zhengming Yu, Zhengyu Lin, et~al.
\newblock Dna-rendering: A diverse neural actor repository for high-fidelity human-centric rendering.
\newblock In \emph{Proceedings of the IEEE/CVF International Conference on Computer Vision}, pages 19982--19993, 2023.

\bibitem[Deng et~al.(2022)Deng, Liu, Zhu, and Ramanan]{dsnerf}
Kangle Deng, Andrew Liu, Jun-Yan Zhu, and Deva Ramanan.
\newblock Depth-supervised nerf: Fewer views and faster training for free.
\newblock In \emph{Proceedings of the IEEE/CVF Conference on Computer Vision and Pattern Recognition}, pages 12882--12891, 2022.

\bibitem[Dong et~al.(2021)Dong, Fang, Jiang, Yang, Bao, and Zhou]{dong2021fast}
Junting Dong, Qi Fang, Wen Jiang, Yurou Yang, Hujun Bao, and Xiaowei Zhou.
\newblock Fast and robust multi-person 3d pose estimation and tracking from multiple views.
\newblock In \emph{T-PAMI}, 2021.

\bibitem[Fridovich-Keil et~al.(2022)Fridovich-Keil, Yu, Tancik, Chen, Recht, and Kanazawa]{fridovich2022plenoxels}
Sara Fridovich-Keil, Alex Yu, Matthew Tancik, Qinhong Chen, Benjamin Recht, and Angjoo Kanazawa.
\newblock Plenoxels: Radiance fields without neural networks.
\newblock In \emph{CVPR}, pages 5501--5510, 2022.

\bibitem[Fridovich-Keil et~al.(2023)Fridovich-Keil, Meanti, Warburg, Recht, and Kanazawa]{fridovich2023k}
Sara Fridovich-Keil, Giacomo Meanti, Frederik~Rahb{\ae}k Warburg, Benjamin Recht, and Angjoo Kanazawa.
\newblock {K-planes: Explicit Radiance Fields in Space, Time, and Appearance}.
\newblock In \emph{Proceedings of the IEEE/CVF Conference on Computer Vision and Pattern Recognition}, pages 12479--12488, 2023.

\bibitem[Gafni et~al.(2021)Gafni, Thies, Zollhofer, and Nie{\ss}ner]{gafni2021dynamic}
Guy Gafni, Justus Thies, Michael Zollhofer, and Matthias Nie{\ss}ner.
\newblock Dynamic neural radiance fields for monocular 4d facial avatar reconstruction.
\newblock In \emph{Proceedings of the IEEE/CVF Conference on Computer Vision and Pattern Recognition}, pages 8649--8658, 2021.

\bibitem[Gan et~al.(2023)Gan, Xu, Huang, Chen, and Yokoya]{gan2023v4d}
Wanshui Gan, Hongbin Xu, Yi Huang, Shifeng Chen, and Naoto Yokoya.
\newblock {V4D: Voxel for 4D Novel View Synthesis}.
\newblock \emph{IEEE Transactions on Visualization and Computer Graphics}, 2023.

\bibitem[Gao et~al.(2021)Gao, Saraf, Kopf, and Huang]{gao2021dynamic}
Chen Gao, Ayush Saraf, Johannes Kopf, and Jia-Bin Huang.
\newblock {Dynamic View Synthesis from Dynamic Monocular Video}.
\newblock In \emph{Proceedings of the IEEE/CVF International Conference on Computer Vision}, pages 5712--5721, 2021.

\bibitem[Gao et~al.(2023)Gao, Wang, Liu, Liu, Theobalt, and Chen]{gao2023neural}
Qingzhe Gao, Yiming Wang, Libin Liu, Lingjie Liu, Christian Theobalt, and Baoquan Chen.
\newblock Neural novel actor: Learning a generalized animatable neural representation for human actors.
\newblock \emph{IEEE Transactions on Visualization and Computer Graphics}, 2023.

\bibitem[Garbin et~al.(2021)Garbin, Kowalski, Johnson, Shotton, and Valentin]{garbin2021fastnerf}
Stephan~J Garbin, Marek Kowalski, Matthew Johnson, Jamie Shotton, and Julien Valentin.
\newblock Fastnerf: High-fidelity neural rendering at 200fps.
\newblock In \emph{ICCV}, pages 14346--14355, 2021.

\bibitem[Geng et~al.(2023)Geng, Peng, Xu, Bao, and Zhou]{geng2023learning}
Chen Geng, Sida Peng, Zhen Xu, Hujun Bao, and Xiaowei Zhou.
\newblock Learning neural volumetric representations of dynamic humans in minutes.
\newblock In \emph{Proceedings of the IEEE/CVF Conference on Computer Vision and Pattern Recognition}, pages 8759--8770, 2023.

\bibitem[Goel et~al.(2023)Goel, Pavlakos, Rajasegaran, Kanazawa, and Malik]{goel2023humans}
Shubham Goel, Georgios Pavlakos, Jathushan Rajasegaran, Angjoo Kanazawa, and Jitendra Malik.
\newblock Humans in 4d: Reconstructing and tracking humans with transformers.
\newblock In \emph{Proceedings of the IEEE/CVF International Conference on Computer Vision}, pages 14783--14794, 2023.

\bibitem[Hu et~al.(2024{\natexlab{a}})Hu, Zhang, Zhang, Zhou, Liu, Zhang, and Nie]{hu2024gaussianavatar}
Liangxiao Hu, Hongwen Zhang, Yuxiang Zhang, Boyao Zhou, Boning Liu, Shengping Zhang, and Liqiang Nie.
\newblock Gaussianavatar: Towards realistic human avatar modeling from a single video via animatable 3d gaussians.
\newblock In \emph{Proceedings of the IEEE/CVF Conference on Computer Vision and Pattern Recognition}, pages 634--644, 2024{\natexlab{a}}.

\bibitem[Hu et~al.(2024{\natexlab{b}})Hu, Hu, and Liu]{hu2024gauhuman}
Shoukang Hu, Tao Hu, and Ziwei Liu.
\newblock Gauhuman: Articulated gaussian splatting from monocular human videos.
\newblock In \emph{Proceedings of the IEEE/CVF Conference on Computer Vision and Pattern Recognition}, pages 20418--20431, 2024{\natexlab{b}}.

\bibitem[Hu et~al.(2024{\natexlab{c}})Hu, Hong, and Liu]{hu2024surmo}
Tao Hu, Fangzhou Hong, and Ziwei Liu.
\newblock Surmo: Surface-based 4d motion modeling for dynamic human rendering.
\newblock In \emph{Proceedings of the IEEE/CVF Conference on Computer Vision and Pattern Recognition}, pages 6550--6560, 2024{\natexlab{c}}.

\bibitem[Jang and Kim(2022)]{jang2022d}
Hankyu Jang and Daeyoung Kim.
\newblock {D-tensorf: Tensorial Radiance Fields for Dynamic Scenes}.
\newblock \emph{arXiv preprint arXiv:2212.02375}, 2022.

\bibitem[Jena et~al.(2023)Jena, Iyer, Choudhary, Smith, Chaudhari, and Gee]{jena2023splatarmor}
Rohit Jena, Ganesh~Subramanian Iyer, Siddharth Choudhary, Brandon Smith, Pratik Chaudhari, and James Gee.
\newblock Splatarmor: Articulated gaussian splatting for animatable humans from monocular rgb videos.
\newblock \emph{arXiv preprint arXiv:2311.10812}, 2023.

\bibitem[Jung et~al.(2023)Jung, Brasch, Song, Perez-Pellitero, Zhou, Li, Navab, and Busam]{jung2023deformable}
HyunJun Jung, Nikolas Brasch, Jifei Song, Eduardo Perez-Pellitero, Yiren Zhou, Zhihao Li, Nassir Navab, and Benjamin Busam.
\newblock Deformable 3d gaussian splatting for animatable human avatars.
\newblock \emph{arXiv preprint arXiv:2312.15059}, 2023.

\bibitem[Kerbl et~al.(2023)Kerbl, Kopanas, Leimk{\"u}hler, and Drettakis]{kerbl3Dgaussians}
Bernhard Kerbl, Georgios Kopanas, Thomas Leimk{\"u}hler, and George Drettakis.
\newblock 3d gaussian splatting for real-time radiance field rendering.
\newblock \emph{ACM Transactions on Graphics}, 42\penalty0 (4), 2023.

\bibitem[Kilian et~al.(2007)Kilian, Mitra, and Pottmann]{kilian2007geometric}
Martin Kilian, Niloy~J Mitra, and Helmut Pottmann.
\newblock Geometric modeling in shape space.
\newblock In \emph{ACM SIGGRAPH 2007 papers}, pages 64--es. 2007.

\bibitem[Kocabas et~al.(2024)Kocabas, Chang, Gabriel, Tuzel, and Ranjan]{kocabas2024hugs}
Muhammed Kocabas, Jen-Hao~Rick Chang, James Gabriel, Oncel Tuzel, and Anurag Ranjan.
\newblock Hugs: Human gaussian splats.
\newblock In \emph{Proceedings of the IEEE/CVF conference on computer vision and pattern recognition}, pages 505--515, 2024.

\bibitem[Kwon et~al.(2021)Kwon, Kim, Ceylan, and Fuchs]{kwon2021neural}
Youngjoong Kwon, Dahun Kim, Duygu Ceylan, and Henry Fuchs.
\newblock Neural human performer: Learning generalizable radiance fields for human performance rendering.
\newblock \emph{Advances in Neural Information Processing Systems}, 34:\penalty0 24741--24752, 2021.

\bibitem[Lei et~al.(2024)Lei, Wang, Pavlakos, Liu, and Daniilidis]{lei2024gart}
Jiahui Lei, Yufu Wang, Georgios Pavlakos, Lingjie Liu, and Kostas Daniilidis.
\newblock Gart: Gaussian articulated template models.
\newblock In \emph{Proceedings of the IEEE/CVF Conference on Computer Vision and Pattern Recognition}, pages 19876--19887, 2024.

\bibitem[Li et~al.(2023)Li, Tao, Yang, and Yang]{li2023human101}
Mingwei Li, Jiachen Tao, Zongxin Yang, and Yi Yang.
\newblock Human101: Training 100+ fps human gaussians in 100s from 1 view.
\newblock \emph{arXiv preprint arXiv:2312.15258}, 2023.

\bibitem[Li et~al.(2024{\natexlab{a}})Li, Yao, Xie, Chen, and Jiang]{li2024gaussianbody}
Mengtian Li, Shengxiang Yao, Zhifeng Xie, Keyu Chen, and Yu-Gang Jiang.
\newblock Gaussianbody: Clothed human reconstruction via 3d gaussian splatting.
\newblock \emph{arXiv preprint arXiv:2401.09720}, 2024{\natexlab{a}}.

\bibitem[Li et~al.(2022)Li, Slavcheva, Zollhoefer, Green, Lassner, Kim, Schmidt, Lovegrove, Goesele, Newcombe, et~al.]{li2022neural}
Tianye Li, Mira Slavcheva, Michael Zollhoefer, Simon Green, Christoph Lassner, Changil Kim, Tanner Schmidt, Steven Lovegrove, Michael Goesele, Richard Newcombe, et~al.
\newblock {Neural 3D Video Synthesis from Multi-view Video}.
\newblock In \emph{Proceedings of the IEEE/CVF Conference on Computer Vision and Pattern Recognition}, pages 5521--5531, 2022.

\bibitem[Li et~al.(2021)Li, Niklaus, Snavely, and Wang]{li2021neural}
Zhengqi Li, Simon Niklaus, Noah Snavely, and Oliver Wang.
\newblock {Neural Scene Flow Fields for Space-time View Synthesis of Dynamic Scenes}.
\newblock In \emph{Proceedings of the IEEE/CVF Conference on Computer Vision and Pattern Recognition}, pages 6498--6508, 2021.

\bibitem[Li et~al.(2024{\natexlab{b}})Li, Zheng, Wang, and Liu]{li2024animatable}
Zhe Li, Zerong Zheng, Lizhen Wang, and Yebin Liu.
\newblock Animatable gaussians: Learning pose-dependent gaussian maps for high-fidelity human avatar modeling.
\newblock In \emph{Proceedings of the IEEE/CVF Conference on Computer Vision and Pattern Recognition}, pages 19711--19722, 2024{\natexlab{b}}.

\bibitem[Lin et~al.(2021)Lin, Ma, Torralba, and Lucey]{lin2021barf}
Chen-Hsuan Lin, Wei-Chiu Ma, Antonio Torralba, and Simon Lucey.
\newblock Barf: Bundle-adjusting neural radiance fields.
\newblock In \emph{ICCV}, pages 5741--5751, 2021.

\bibitem[Lin et~al.(2023)Lin, Peng, Xu, Xie, He, Bao, and Zhou]{lin2023im4d}
Haotong Lin, Sida Peng, Zhen Xu, Tao Xie, Xingyi He, Hujun Bao, and Xiaowei Zhou.
\newblock High-fidelity and real-time novel view synthesis for dynamic scenes.
\newblock In \emph{SIGGRAPH Asia Conference Proceedings}, 2023.

\bibitem[Liu et~al.(2020)Liu, Gu, Zaw~Lin, Chua, and Theobalt]{liu2020neural}
Lingjie Liu, Jiatao Gu, Kyaw Zaw~Lin, Tat-Seng Chua, and Christian Theobalt.
\newblock Neural sparse voxel fields.
\newblock pages 15651--15663, 2020.

\bibitem[Liu et~al.(2024)Liu, Wu, Liu, Liu, Wu, Zhao, Feng, Ding, and Wang]{liu2024gea}
Xinqi Liu, Chenming Wu, Xing Liu, Jialun Liu, Jinbo Wu, Chen Zhao, Haocheng Feng, Errui Ding, and Jingdong Wang.
\newblock Gea: Reconstructing expressive 3d gaussian avatar from monocular video.
\newblock \emph{arXiv preprint arXiv:2402.16607}, 2024.

\bibitem[Liu et~al.(2023{\natexlab{a}})Liu, Huang, Qin, Lin, and Wang]{liu2023animatable}
Yang Liu, Xiang Huang, Minghan Qin, Qinwei Lin, and Haoqian Wang.
\newblock Animatable 3d gaussian: Fast and high-quality reconstruction of multiple human avatars.
\newblock \emph{arXiv preprint arXiv:2311.16482}, 2023{\natexlab{a}}.

\bibitem[Liu et~al.(2023{\natexlab{b}})Liu, Gao, Meuleman, Tseng, Saraf, Kim, Chuang, Kopf, and Huang]{liu2023robust}
Yu-Lun Liu, Chen Gao, Andreas Meuleman, Hung-Yu Tseng, Ayush Saraf, Changil Kim, Yung-Yu Chuang, Johannes Kopf, and Jia-Bin Huang.
\newblock {Robust Dynamic Radiance Fields}.
\newblock In \emph{Proceedings of the IEEE/CVF Conference on Computer Vision and Pattern Recognition}, pages 13--23, 2023{\natexlab{b}}.

\bibitem[Loper et~al.(2015)Loper, Mahmood, Romero, Pons-Moll, and Black]{loper2015smpl}
Matthew Loper, Naureen Mahmood, Javier Romero, Gerard Pons-Moll, and Michael~J Black.
\newblock Smpl: A skinned multi-person linear model.
\newblock \emph{ACM Transactions on Graphics}, 34\penalty0 (6), 2015.

\bibitem[Mildenhall et~al.(2020)Mildenhall, Srinivasan, Tancik, Barron, Ramamoorthi, and Ng]{mildenhall2020nerf}
Ben Mildenhall, Pratul~P Srinivasan, Matthew Tancik, Jonathan~T Barron, Ravi Ramamoorthi, and Ren Ng.
\newblock Nerf: Representing scenes as neural radiance fields for view synthesis.
\newblock In \emph{European Conference on Computer Vision}, pages 405--421, 2020.

\bibitem[Moreau et~al.(2024)Moreau, Song, Dhamo, Shaw, Zhou, and P{\'e}rez-Pellitero]{moreau2024human}
Arthur Moreau, Jifei Song, Helisa Dhamo, Richard Shaw, Yiren Zhou, and Eduardo P{\'e}rez-Pellitero.
\newblock Human gaussian splatting: Real-time rendering of animatable avatars.
\newblock In \emph{Proceedings of the IEEE/CVF Conference on Computer Vision and Pattern Recognition}, pages 788--798, 2024.

\bibitem[M{\"u}ller et~al.(2022)M{\"u}ller, Evans, Schied, and Keller]{muller2022instant}
Thomas M{\"u}ller, Alex Evans, Christoph Schied, and Alexander Keller.
\newblock Instant neural graphics primitives with a multiresolution hash encoding.
\newblock \emph{ACM Transactions on Graphics (ToG)}, 41\penalty0 (4):\penalty0 1--15, 2022.

\bibitem[Niu et~al.(2024)Niu, Zhan, Zhu, Li, Wang, Zhong, Sun, and Zheng]{niu2024bundle}
Muyao Niu, Yifan Zhan, Qingtian Zhu, Zhuoxiao Li, Wei Wang, Zhihang Zhong, Xiao Sun, and Yinqiang Zheng.
\newblock Bundle adjusted gaussian avatars deblurring.
\newblock \emph{arXiv preprint arXiv:2411.16758}, 2024.

\bibitem[Niu et~al.(2025)Niu, Cao, Zhan, Zhu, Ma, Zhao, Zeng, Zhong, Sun, and Zheng]{niu2025anicrafter}
Muyao Niu, Mingdeng Cao, Yifan Zhan, Qingtian Zhu, Mingze Ma, Jiancheng Zhao, Yanhong Zeng, Zhihang Zhong, Xiao Sun, and Yinqiang Zheng.
\newblock Anicrafter: Customizing realistic human-centric animation via avatar-background conditioning in video diffusion models.
\newblock \emph{arXiv preprint arXiv:2505.20255}, 2025.

\bibitem[Park et~al.(2021)Park, Sinha, Barron, Bouaziz, Goldman, Seitz, and Martin-Brualla]{park2021nerfies}
Keunhong Park, Utkarsh Sinha, Jonathan~T Barron, Sofien Bouaziz, Dan~B Goldman, Steven~M Seitz, and Ricardo Martin-Brualla.
\newblock {Nerfies: Deformable Neural Radiance Fields}.
\newblock In \emph{Proceedings of the IEEE/CVF International Conference on Computer Vision}, pages 5865--5874, 2021.

\bibitem[Park et~al.(2023)Park, Son, Jang, Ahn, Kim, and Kang]{park2023temporal}
Sungheon Park, Minjung Son, Seokhwan Jang, Young~Chun Ahn, Ji-Yeon Kim, and Nahyup Kang.
\newblock {Temporal Interpolation Is All You Need for Dynamic Neural Radiance Fields}.
\newblock In \emph{Proceedings of the IEEE/CVF Conference on Computer Vision and Pattern Recognition}, pages 4212--4221, 2023.

\bibitem[Pavlakos et~al.(2019)Pavlakos, Choutas, Ghorbani, Bolkart, Osman, Tzionas, and Black]{SMPL-X:2019}
Georgios Pavlakos, Vasileios Choutas, Nima Ghorbani, Timo Bolkart, Ahmed A.~A. Osman, Dimitrios Tzionas, and Michael~J. Black.
\newblock Expressive body capture: {3D} hands, face, and body from a single image.
\newblock In \emph{Proceedings IEEE Conf. on Computer Vision and Pattern Recognition (CVPR)}, pages 10975--10985, 2019.

\bibitem[Peng et~al.(2024)Peng, Xu, Tang, Jiao, Wang, et~al.]{peng2024structure}
Rui Peng, Wangze Xu, Luyang Tang, Jianbo Jiao, Ronggang Wang, et~al.
\newblock Structure consistent gaussian splatting with matching prior for few-shot novel view synthesis.
\newblock \emph{Advances in Neural Information Processing Systems}, 37:\penalty0 97328--97352, 2024.

\bibitem[Peng et~al.(2021)Peng, Zhang, Xu, Wang, Shuai, Bao, and Zhou]{peng2021neural}
Sida Peng, Yuanqing Zhang, Yinghao Xu, Qianqian Wang, Qing Shuai, Hujun Bao, and Xiaowei Zhou.
\newblock Neural body: Implicit neural representations with structured latent codes for novel view synthesis of dynamic humans.
\newblock In \emph{CVPR}, 2021.

\bibitem[Prokudin et~al.(2023)Prokudin, Ma, Raafat, Valentin, and Tang]{prokudin2023dynamic}
Sergey Prokudin, Qianli Ma, Maxime Raafat, Julien Valentin, and Siyu Tang.
\newblock Dynamic point fields.
\newblock In \emph{Proceedings of the IEEE/CVF International Conference on Computer Vision}, pages 7964--7976, 2023.

\bibitem[Pumarola et~al.(2021)Pumarola, Corona, Pons-Moll, and Moreno-Noguer]{pumarola2021d}
Albert Pumarola, Enric Corona, Gerard Pons-Moll, and Francesc Moreno-Noguer.
\newblock {D-neRF: Neural Radiance Fields for Dynamic Scenes}.
\newblock In \emph{Proceedings of the IEEE/CVF Conference on Computer Vision and Pattern Recognition}, pages 10318--10327, 2021.

\bibitem[Qian et~al.(2024)Qian, Wang, Mihajlovic, Geiger, and Tang]{qian20243dgs}
Zhiyin Qian, Shaofei Wang, Marko Mihajlovic, Andreas Geiger, and Siyu Tang.
\newblock 3dgs-avatar: Animatable avatars via deformable 3d gaussian splatting.
\newblock In \emph{Proceedings of the IEEE/CVF Conference on Computer Vision and Pattern Recognition}, pages 5020--5030, 2024.

\bibitem[Shao et~al.(2023)Shao, Zheng, Tu, Liu, Zhang, and Liu]{shao2023tensor4d}
Ruizhi Shao, Zerong Zheng, Hanzhang Tu, Boning Liu, Hongwen Zhang, and Yebin Liu.
\newblock {Tensor4d: Efficient Neural 4D Decomposition for High-fidelity Dynamic Reconstruction and Rendering}.
\newblock In \emph{Proceedings of the IEEE/CVF Conference on Computer Vision and Pattern Recognition}, pages 16632--16642, 2023.

\bibitem[Shen et~al.(2024)Shen, Gao, Xu, Peng, Tang, Xiong, Jiao, and Wang]{shen2024disentangled}
Shihe Shen, Huachen Gao, Wangze Xu, Rui Peng, Luyang Tang, Kaiqiang Xiong, Jianbo Jiao, and Ronggang Wang.
\newblock Disentangled generation and aggregation for robust radiance fields.
\newblock In \emph{European Conference on Computer Vision}, pages 218--236. Springer, 2024.

\bibitem[Shuai et~al.(2022)Shuai, Geng, Fang, Peng, Shen, Zhou, and Bao]{shuai2022multinb}
Qing Shuai, Chen Geng, Qi Fang, Sida Peng, Wenhao Shen, Xiaowei Zhou, and Hujun Bao.
\newblock Novel view synthesis of human interactions from sparse multi-view videos.
\newblock In \emph{SIGGRAPH Conference Proceedings}, 2022.

\bibitem[Sun et~al.(2022)Sun, Sun, and Chen]{sun2022direct}
Cheng Sun, Min Sun, and Hwann-Tzong Chen.
\newblock Direct voxel grid optimization: Super-fast convergence for radiance fields reconstruction.
\newblock In \emph{CVPR}, pages 5459--5469, 2022.

\bibitem[Truong et~al.(2023)Truong, Rakotosaona, Manhardt, and Tombari]{truong2023sparf}
Prune Truong, Marie-Julie Rakotosaona, Fabian Manhardt, and Federico Tombari.
\newblock Sparf: Neural radiance fields from sparse and noisy poses.
\newblock In \emph{Proceedings of the IEEE/CVF Conference on Computer Vision and Pattern Recognition}, pages 4190--4200, 2023.

\bibitem[Tu et~al.(2025)Tu, Xing, Han, Cheng, Dai, Luo, and Wu]{tu2025stableanimator}
Shuyuan Tu, Zhen Xing, Xintong Han, Zhi-Qi Cheng, Qi Dai, Chong Luo, and Zuxuan Wu.
\newblock Stableanimator: High-quality identity-preserving human image animation.
\newblock In \emph{Proceedings of the Computer Vision and Pattern Recognition Conference}, pages 21096--21106, 2025.

\bibitem[Wang et~al.(2023)Wang, Chen, Loy, and Liu]{wang2023sparsenerf}
Guangcong Wang, Zhaoxi Chen, Chen~Change Loy, and Ziwei Liu.
\newblock Sparsenerf: Distilling depth ranking for few-shot novel view synthesis.
\newblock In \emph{Proceedings of the IEEE/CVF international conference on computer vision}, pages 9065--9076, 2023.

\bibitem[Wang et~al.(2022)Wang, Schwarz, Geiger, and Tang]{wang2022arah}
Shaofei Wang, Katja Schwarz, Andreas Geiger, and Siyu Tang.
\newblock Arah: Animatable volume rendering of articulated human sdfs.
\newblock In \emph{European conference on computer vision}, pages 1--19. Springer, 2022.

\bibitem[Wang et~al.(2024)Wang, Zhang, Gao, Wang, Zhou, Zhang, Yan, and Sang]{wang2024unianimate}
Xiang Wang, Shiwei Zhang, Changxin Gao, Jiayu Wang, Xiaoqiang Zhou, Yingya Zhang, Luxin Yan, and Nong Sang.
\newblock Unianimate: Taming unified video diffusion models for consistent human image animation.
\newblock \emph{arXiv preprint arXiv:2406.01188}, 2024.

\bibitem[Wang et~al.(2004)Wang, Bovik, Sheikh, and Simoncelli]{wang2004image}
Zhou Wang, Alan~C Bovik, Hamid~R Sheikh, and Eero~P Simoncelli.
\newblock {Image Quality Assessment: from Error Visibility to Structural Similarity}.
\newblock \emph{IEEE transactions on image processing}, 13\penalty0 (4):\penalty0 600--612, 2004.

\bibitem[Wang et~al.(2021)Wang, Wu, Xie, Chen, and Prisacariu]{wang2021nerf}
Zirui Wang, Shangzhe Wu, Weidi Xie, Min Chen, and Victor~Adrian Prisacariu.
\newblock Nerf--: Neural radiance fields without known camera parameters.
\newblock \emph{arXiv preprint arXiv:2102.07064}, 2021.

\bibitem[Weng et~al.(2022)Weng, Curless, Srinivasan, Barron, and Kemelmacher-Shlizerman]{weng2022humannerf}
Chung-Yi Weng, Brian Curless, Pratul~P Srinivasan, Jonathan~T Barron, and Ira Kemelmacher-Shlizerman.
\newblock Humannerf: Free-viewpoint rendering of moving people from monocular video.
\newblock In \emph{Proceedings of the IEEE/CVF conference on computer vision and pattern Recognition}, pages 16210--16220, 2022.

\bibitem[Wu et~al.(2025)Wu, Peng, Wang, Xiao, Tang, Yan, Xiong, and Wang]{wu2025swift4d}
Jiahao Wu, Rui Peng, Zhiyan Wang, Lu Xiao, Luyang Tang, Jinbo Yan, Kaiqiang Xiong, and Ronggang Wang.
\newblock Swift4d: Adaptive divide-and-conquer gaussian splatting for compact and efficient reconstruction of dynamic scene.
\newblock \emph{arXiv preprint arXiv:2503.12307}, 2025.

\bibitem[Xian et~al.(2021)Xian, Huang, Kopf, and Kim]{xian2021space}
Wenqi Xian, Jia-Bin Huang, Johannes Kopf, and Changil Kim.
\newblock {Space-time Neural Irradiance Fields for Free-viewpoint Video}.
\newblock In \emph{Proceedings of the IEEE/CVF Conference on Computer Vision and Pattern Recognition}, pages 9421--9431, 2021.

\bibitem[Xu et~al.(2024{\natexlab{a}})Xu, Gao, Shen, Peng, Jiao, and Wang]{xu2024mvpgs}
Wangze Xu, Huachen Gao, Shihe Shen, Rui Peng, Jianbo Jiao, and Ronggang Wang.
\newblock Mvpgs: Excavating multi-view priors for gaussian splatting from sparse input views.
\newblock In \emph{European Conference on Computer Vision}, pages 203--220. Springer, 2024{\natexlab{a}}.

\bibitem[Xu et~al.(2024{\natexlab{b}})Xu, Peng, Lin, He, Sun, Shen, Bao, and Zhou]{xu20244k4d}
Zhen Xu, Sida Peng, Haotong Lin, Guangzhao He, Jiaming Sun, Yujun Shen, Hujun Bao, and Xiaowei Zhou.
\newblock 4k4d: Real-time 4d view synthesis at 4k resolution.
\newblock In \emph{CVPR}, 2024{\natexlab{b}}.

\bibitem[Xu et~al.(2024{\natexlab{c}})Xu, Zhang, Liew, Yan, Liu, Zhang, Feng, and Shou]{xu2024magicanimate}
Zhongcong Xu, Jianfeng Zhang, Jun~Hao Liew, Hanshu Yan, Jia-Wei Liu, Chenxu Zhang, Jiashi Feng, and Mike~Zheng Shou.
\newblock Magicanimate: Temporally consistent human image animation using diffusion model.
\newblock In \emph{Proceedings of the IEEE/CVF Conference on Computer Vision and Pattern Recognition}, pages 1481--1490, 2024{\natexlab{c}}.

\bibitem[Yan et~al.(2025)Yan, Peng, Wang, Tang, Yang, Liang, Wu, and Wang]{yan2025instant}
Jinbo Yan, Rui Peng, Zhiyan Wang, Luyang Tang, Jiayu Yang, Jie Liang, Jiahao Wu, and Ronggang Wang.
\newblock Instant gaussian stream: Fast and generalizable streaming of dynamic scene reconstruction via gaussian splatting.
\newblock In \emph{Proceedings of the Computer Vision and Pattern Recognition Conference}, pages 16520--16531, 2025.

\bibitem[Zhan et~al.(2024)Zhan, Zhu, Niu, Ma, Zhao, Zhong, Sun, Qiao, and Zheng]{zhan2024tomie}
Yifan Zhan, Qingtian Zhu, Muyao Niu, Mingze Ma, Jiancheng Zhao, Zhihang Zhong, Xiao Sun, Yu Qiao, and Yinqiang Zheng.
\newblock Tomie: Towards modular growth in enhanced smpl skeleton for 3d human with animatable garments.
\newblock \emph{arXiv preprint arXiv:2410.08082}, 2024.

\bibitem[Zhan et~al.(2025)Zhan, Xu, Zhu, Niu, Ma, Liu, Zhong, Sun, and Zheng]{zhan2025r3}
Yifan Zhan, Wangze Xu, Qingtian Zhu, Muyao Niu, Mingze Ma, Yifei Liu, Zhihang Zhong, Xiao Sun, and Yinqiang Zheng.
\newblock R3-avatar: Record and retrieve temporal codebook for reconstructing photorealistic human avatars.
\newblock \emph{arXiv preprint arXiv:2503.12751}, 2025.

\bibitem[Zhang et~al.(2018)Zhang, Isola, Efros, Shechtman, and Wang]{zhang2018unreasonable}
Richard Zhang, Phillip Isola, Alexei~A Efros, Eli Shechtman, and Oliver Wang.
\newblock {The Unreasonable Effectiveness of Deep Features as a Perceptual Metric}.
\newblock In \emph{Proceedings of the IEEE conference on computer vision and pattern recognition}, pages 586--595, 2018.

\bibitem[Zheng et~al.(2024{\natexlab{a}})Zheng, Zhou, Shao, Liu, Zhang, Nie, and Liu]{zheng2024gps}
Shunyuan Zheng, Boyao Zhou, Ruizhi Shao, Boning Liu, Shengping Zhang, Liqiang Nie, and Yebin Liu.
\newblock Gps-gaussian: Generalizable pixel-wise 3d gaussian splatting for real-time human novel view synthesis.
\newblock In \emph{Proceedings of the IEEE/CVF Conference on Computer Vision and Pattern Recognition}, pages 19680--19690, 2024{\natexlab{a}}.

\bibitem[Zheng et~al.(2024{\natexlab{b}})Zheng, Zhao, Yang, Yifan, Xiang, Dubost, Lagun, Beeler, Tombari, Guibas, et~al.]{zheng2024physavatar}
Yang Zheng, Qingqing Zhao, Guandao Yang, Wang Yifan, Donglai Xiang, Florian Dubost, Dmitry Lagun, Thabo Beeler, Federico Tombari, Leonidas Guibas, et~al.
\newblock Physavatar: Learning the physics of dressed 3d avatars from visual observations.
\newblock \emph{arXiv preprint arXiv:2404.04421}, 2024{\natexlab{b}}.

\bibitem[Zhu et~al.(2024)Zhu, Fan, Jiang, and Wang]{zhu2024fsgs}
Zehao Zhu, Zhiwen Fan, Yifan Jiang, and Zhangyang Wang.
\newblock Fsgs: Real-time few-shot view synthesis using gaussian splatting.
\newblock In \emph{European conference on computer vision}, pages 145--163. Springer, 2024.

\bibitem[Zielonka et~al.(2023)Zielonka, Bagautdinov, Saito, Zollh{\"o}fer, Thies, and Romero]{zielonka2023drivable}
Wojciech Zielonka, Timur Bagautdinov, Shunsuke Saito, Michael Zollh{\"o}fer, Justus Thies, and Javier Romero.
\newblock Drivable 3d gaussian avatars.
\newblock \emph{arXiv preprint arXiv:2311.08581}, 2023.

\end{thebibliography}
